\documentclass{article}

\usepackage{ijcai23}

\usepackage{times}
\usepackage[utf8]{inputenc}
\usepackage{soul}
\usepackage{amsmath,amssymb,amsfonts,amsthm,bm,mathtools}
\allowdisplaybreaks
\usepackage{graphicx,subfigure}
\usepackage{booktabs}

\usepackage{multirow}
\usepackage{makecell}
\usepackage[small]{caption}
\usepackage{enumitem}
\usepackage{url}
\usepackage[hidelinks]{hyperref}
\usepackage{natbib}
\usepackage{xcolor}
\usepackage[switch]{lineno}

\newcommand{\bB}{\mathbf{B}}

\newcommand{\bD}{\mathbf{D}}

\newcommand{\bM}{\mathbf{M}}

\newcommand{\bX}{\mathbf{X}}
\newcommand{\bY}{\mathbf{Y}}

\newcommand{\bu}{\mathbf{u}}
\newcommand{\bv}{\mathbf{v}}
\newcommand{\bw}{\mathbf{w}}
\newcommand{\bx}{\mathbf{x}}
\newcommand{\by}{\mathbf{y}}

\newcommand{\cD}{\mathcal{D}}

\newcommand{\cH}{\mathcal{H}}

\newcommand{\cN}{\mathcal{N}}

\newcommand{\E}{\mathbb{E}}

\newcommand{\R}{\mathbb{R}}
\newcommand{\btheta}{\bm{\theta}}

\newcommand{\bphi}{\bm{\phi}}

\newcommand{\Std}{\mathrm{Std}}
\newcommand{\Var}{\mathrm{Var}}
\newcommand{\Cov}{\mathrm{Cov}}

\newtheorem{assumption}{Assumption}

\newtheorem{theorem}{Theorem}

\newcommand{\specialcell}[2][c]{\begin{tabular}[#1]{@{}c@{}}#2\end{tabular}}

\newcommand{\ours}{StableMiss}

\title{Prediction with Incomplete Data under Agnostic Mask Distribution Shift}

\author{
Yichen~Zhu$^1$\and
Jian~Yuan$^1$\and
Bo~Jiang\footnote{Corresponding author.}$^{1}$\and
Tao~Lin$^2$\and
Haiming~Jin$^1$\and
Xinbing~Wang$^1$\And
Chenghu~Zhou$^1$\\
\affiliations
$^1$Shanghai Jiao Tong University, $^2$Communication University of China\\
\emails
\{zyc\_ieee, yuanjian, bjiang\}@sjtu.edu.cn, lintao@cuc.edu.cn, \{jinhaiming, xwang8\}@sjtu.edu.cn, zhouchsjtu@gmail.com\\
}

\begin{document}

\maketitle

\begin{abstract}
Data with missing values is ubiquitous in many applications. Recent years have witnessed increasing attention on prediction with only incomplete data consisting of observed features and a mask that indicates the missing pattern. Existing methods assume that the training and testing distributions are the same, which may be violated in real-world scenarios.
In this paper, we consider prediction with incomplete data in the presence of distribution shift. We focus on the case where the underlying joint distribution of complete features and label is invariant, but the missing pattern, i.e., mask distribution may shift agnostically between training and testing.
To achieve generalization, we leverage the observation that for each mask, there is an invariant optimal predictor. To avoid the exponential explosion when learning them separately, we approximate the optimal predictors jointly using a double parameterization technique. This has the undesirable side effect of allowing the learned predictors to rely on the intra-mask correlation and that between features and mask. We perform decorrelation to minimize this effect. Combining
the techniques above, we propose a novel prediction method called {\ours}. Extensive experiments on both synthetic and real-world datasets show that {\ours} is robust and outperforms state-of-the-art methods under agnostic mask distribution shift.
\end{abstract}


\section{Introduction}\label{sec:introduction}

Data with missing values is ubiquitous in many applications due to sensor malfunction, incomplete sensing coverage, etc. Recent years have witnessed increasing attention on prediction with only incomplete feature, which consists of observed feature values and a mask that indicates which features are observed. Existing methods \citep{morvan2020neumiss,morvan2021whats} assume that the training and testing distributions are the same. However, this assumption can be violated in real-world scenarios. In this paper, we study the problem of prediction with incomplete data in the presence of distribution shift. We focus on the scenario where the underlying joint distribution of complete features and label is invariant, but the missing pattern, i.e., mask distribution may be different between training and testing. Such mask distribution shift may result from different sensor deployment, data management, etc. For example, DiDi Traffic Speed \citep{didi2018city} is a naturally incomplete dataset throughout one year. The traffic network is almost unchanged, so we may reasonably assume the speed distribution is relatively stable. The average missing rate from Jan to Jun is 37\% but drops to 23\% from Jul to Dec, possibly because more and higher-quality sensors are deployed. Moreover, different from the transfer learning formulation \citep{pan2010a}, we assume the mask distribution shift is agnostic, since testing distribution is usually unavailable during training in practice.

Many methods for missing data can be used to predict with incomplete feature. \citet{morvan2020neumiss} proposes the NeuMiss network for linear regression with incomplete Gaussian feature. \citet{morvan2021whats} extends NeuMiss to nonlinear case by training it jointly with an MLP. Some missing data imputation methods \citep{yoon2018gain,li2019learning,ma2019eddi,mattei2019miwae,li2020learning} can also be used by treating the label as missing feature. However, none of these methods consider distribution shift. Their models learn the information of mask distribution and thus can hardly generalize under mask distribution shift.

Several methods have been proposed for prediction under agnostic feature distribution shift \citep{shen2018causally,kuang2018stable,kuang2020stable,shen2020stable,zhang2021deep,xu2022a}. It is assumed that the conditional label distribution given complete feature is invariant. They learn this invariant conditional distribution to achieve generalization. However, they are designed for complete data. In our setting, the conditional label distribution given trivially-imputed complete feature is not invariant, so these methods cannot be applied to incomplete data.

We observe that the conditional label distribution given observed feature values and mask is invariant between training and testing. As a result, there is an invariant optimal predictor for each mask respectively. Learning these optimal predictors can generalize under agnostic mask distribution shift. Since the number of optimal predictors increases exponentially with feature dimension \citep{morvan2020neumiss}, we approximate the optimal predictors jointly using a double parameterization technique, i.e., we first parameterize the optimal predictors and then further parameterize the mapping from mask to the parameters of the optimal predictors. However, such parameterization has the undesirable side effect that the learned model may depend on the intra-mask correlation and the correlation between features and mask, since the training loss depends on these correlations. This may defy the generalizability of the learned model, as such correlations will change under mask distribution shift. Inspired by \citet{xu2022a}, we decorrelate features and mask to make the learned model independent of their correlations and help it well approximate the optimal predictors. Combining all the techniques above, we have our {\ours} that can achieve generalization under agnostic mask distribution shift.

The contributions of this paper are summarized as follows.
\begin{itemize}[leftmargin=*]
	\item This paper proposes {\ours}, a novel method for prediction with incomplete data  that is robust to agnostic mask distribution shift. To the best of our knowledge, this is the first method that considers agnostic mask distribution shift.
	\item Extensive experiments are conducted on both synthetic and real-world datasets. The results show that {\ours} is robust and outperforms the state-of-the-art methods under agnostic mask distribution shift.
\end{itemize}


\section{Related Work}\label{sec:related_work}

\textbf{Prediction with Incomplete Data}. Prediction with incomplete feature has attracted increasing attention recently. \citep{morvan2020neumiss} derives the analytical expression of the optimal predictor for linear regression with incomplete Gaussian feature and proposes the NeuMiss network to approximate the optimal predictor. \citep{morvan2021whats} then extends NeuMiss to nonlinear case by training it jointly with a Multi-Layer Perceptron (MLP). Besides these methods natively proposed for prediction, some missing data imputation methods can also be used by treating the label as missing feature. GAIN \citep{yoon2018gain} is an adaptation of GAN \citep{goodfellow2014generative}, where the generator imputes missing entries, which the discriminator tries to distinguish from observed entries with partial information about the mask. MisGAN \citep{li2019learning} learns the complete data distribution from incomplete data and uses it to supervise imputation. Partial VAE \citep{ma2019eddi}, MIWAE \citep{mattei2019miwae} and P-BiGAN \citep{li2020learning} extend VAE \citep{kingma2014auto-encoding}, IWAE \citep{burda2016importance} and BiGAN \citep{donahue2017adversarial} respectively to learn the prior and posterior distributions of incomplete feature given its latent representation and mask. However, all these methods assume that the training and testing distributions are the same. They can hardly generalize under distribution shift.

\textbf{Prediction under Agnostic Feature Distribution Shift}. Several methods have been proposed for prediction under agnostic feature distribution shift. They typically assume that the conditional label distribution given complete feature is invariant. Then they learn this invariant conditional distribution to achieve generalization by decorrelating the features. CRLR \citep{shen2018causally} learns a weight for each sample respectively by minimizing the so-called confounder balancing loss, which is zero when the features are decorrelated. Then weighted Logistic regression is carried out with the learned weights. Based on CRLR, DGBR \citep{kuang2018stable} also extracts nonlinear representations of features with deep auto-encoder. DWR \citep{kuang2020stable} learns a set of sample weights by minimizing the sum of covariance between pairs of features and then carries out weighted linear regression. SRDO \citep{shen2020stable} first constructs a new dataset by sampling each feature independently from the training set to decorrelate among features and then learns sample weights by density ratio estimation \citep{sugiyama2012density}. Weighted linear regression is also used with the learned weights. StableNet \citep{zhang2021deep} adopts Random Fourier Features to measure nonlinear correlations among features. It iteratively optimizes a weighted regression model and a set of sample weights by minimizing the prediction error and the nonlinear correlations, respectively. \citet{xu2022a} proposes a general framework with DWR and SRDO as specific implementations and gives theoretical analysis on the framework. However, all these methods are designed for complete data and cannot be applied to incomplete data.

Compared to the existing methods, {\ours} can not only predict with incomplete feature but also generalize under agnostic mask distribution shift.


\section{Problem Formulation}\label{sec:problem_formulation}

\subsection{Preliminary}\label{sec:preliminary}

We use capital and lowercase letters, e.g., $X$ and $x$, to denote random variable and its realization, respectively. We use  subscripts to index the entries of a vector, e.g., $x_i$ is the $i$-th entry of $\bx$. Let $\bx\in\R^n$ and $\by\in\R^d$ denote the feature and label, respectively. In the presence of missing data, we consider the case where $\bx$ is partially observed and $\by$ is fully observed during training. A binary mask $\mathbf{m}\in\{0,1\}^n$ indicates which entries of $\bx$ are observed: $m_i=1$ if $x_i$ is observed, and $m_i=0$ if $x_i$ is missing. The complementary mask $\overline{\mathbf{m}}$ is defined by $\overline{m}_i=1-m_i$, $\forall i$. With a slight abuse of notation, we regard $\mathbf{m}$ and $\overline{\mathbf{m}}$ as the index sets of the observed and missing entries, so that the observed and missing feature values are $\bx_{\mathbf{m}}=\{x_i\mid i\in\mathbf{m}\}$ and $\bx_{\overline{\mathbf{m}}}=\{x_i\mid i\in\overline{\mathbf{m}}\}$, respectively. We consider the case where the mask $\mathbf{m}$ is known, since it is common to know which features are observed within incomplete feature. The incomplete feature is given by $(\bx_{\mathbf{m}},\mathbf{m})$. We consider the case where label generation process depends on the feature but not the mask, i.e., $p(\by\mid\bx,\mathbf{m})=p(\by\mid\bx)$. We do not make assumptions on $p(\bx)$.

Following \citet{little1986statistical}, we model the generative process of incomplete feature as follows. A complete feature sample $\bx$ is first drawn from the complete feature distribution $p(\bx)$. Given $\bx$, a mask sample $\mathbf{m}$ is then drawn from the conditional mask distribution $p(\mathbf{m}\mid\bx)$. The resulted incomplete feature $(\bx_{\mathbf{m}},\mathbf{m})$ follows the distribution
\begin{equation*}
	p(\bx_{\mathbf{m}},\mathbf{m})=\int p(\bx)p(\mathbf{m}\mid\bx)d\bx_{\overline{\mathbf{m}}}\text{.}
\end{equation*}
We focus on the Missing Completely At Random (MCAR) case and Missing At Random (MAR) case \citep{little1986statistical}. Under MCAR, the mask $\bM$ is independent of the underlying complete feature $\bX$, i.e., $p(\mathbf{m}\mid\bx)=p(\mathbf{m}), \forall \mathbf{m}, \bx$; under MAR, $\bM$ only depends on the observed feature values $\bX_\bM$, i.e., $p(\mathbf{m}\mid\bx)=p(\mathbf{m}\mid\bx_{\mathbf{m}}), \forall \mathbf{m}, \bx$.

\subsection{Problem Statement}\label{sec:problem_statement}

The problem is to predict with incomplete feature under agnostic mask distribution shift. Given a training set $\cD=\{(\bx_{\mathbf{m}^{(i)}}^{(i)},\mathbf{m}^{(i)},\by^{(i)})\}_{i=1}^N$, consisting of $N$ samples from the training distribution $p^{tr}(\bx,\mathbf{m},\by)$, the goal is to learn a prediction function $g(\bx_{\mathbf{m}},\mathbf{m})$ for agnostic testing distribution $p^{te}(\bx,\mathbf{m},\by)$, where the input to $g$ is only incomplete feature. We seek the prediction function $g$ to minimize the mean squared loss
\begin{equation*}
    \ell(g)=\E_{(\bX,\bM,\bY)\sim p^{te}}\|\bY-g(\bX_\bM,\bM)\|_2^2\text{.}
\end{equation*}
Ideally, the optimal $g$ is given by the conditional expectation:
\begin{equation*}
g(\bx_{\mathbf{m}}, \mathbf{m})=\E_{\bY\sim p_{\bY\mid\bx_{\mathbf{m}},\mathbf{m}}^{te}}[\bY\mid \bx_\mathbf{m},\mathbf{m}].
\end{equation*}
How we learn it approximately from training data will be introduced in the next section.

Note that the testing error can be arbitrarily large without any prior knowledge about the testing distribution. We consider the case where only the mask distribution may change between training and testing. More specifically, we make the following assumption on the testing distribution.

\begin{assumption}\label{assump:invariant_joint_distribution}
	The joint distribution of complete feature and label is invariant between training and testing:
	\begin{equation*}
		p^{te}(\bx,\by)=p^{tr}(\bx,\by)\text{.}
	\end{equation*}
\end{assumption}

The mask distribution shift still remains agnostic under the above assumption. Different from the transfer learning formulation \citep{pan2010a}, the testing distribution is unavailable during the training process.


\section{Methodology}\label{sec:methodology}

\subsection{Prediction Framework}\label{sec:prediction_framework}

Our method relies critically on the following simple result, the proof of which is given in Appendix \ref{app:proof_invariant_optimal_predictor}.

\begin{theorem}\label{thm:invariant_optimal_predictor}
  Under Assumption \ref{assump:invariant_joint_distribution}, the conditional label distribution given observed feature values and mask is invariant between training and testing in MCAR or MAR:
	\begin{equation*}
	 p^{te}(\by\mid\bx_\mathbf{m},\mathbf{m})=p^{tr}(\by\mid\bx_\mathbf{m},\mathbf{m})\text{.}
	\end{equation*}
  As a consequence, 
  \begin{equation*}
    \E_{\bY\sim p_{\bY\mid\bx_{\mathbf{m}},\mathbf{m}}^{te}}[\bY\mid\bx_\mathbf{m},\mathbf{m}]=\E_{\bY\sim p_{\bY\mid\bx_{\mathbf{m}},\mathbf{m}}^{tr}}[\bY\mid\bx_\mathbf{m},\mathbf{m}]\text{.}
  \end{equation*}
\end{theorem}

Theorem \ref{thm:invariant_optimal_predictor} holds since MCAR and MAR guarantee that the missing feature $\bX_{\overline{\bM}}$ is independent of of mask $\bM$ given the observed feature $\bX_\bM$. It shows that the ideal optimal predictor is invariant between training and testing. If we can learn it under the training distribution, it will automatically generalize to the agnostic testing distribution. However, as noted in \citet{morvan2021whats}, it is essentially an aggregation of $2^n$ optimal predictors, one for each specific mask $\mathbf{m}$. Since the number of optimal predictors increases exponentially with feature dimension, it is infeasible to learn them separately. 

\begin{figure}[t]
	\centering
	\includegraphics[width=\columnwidth]{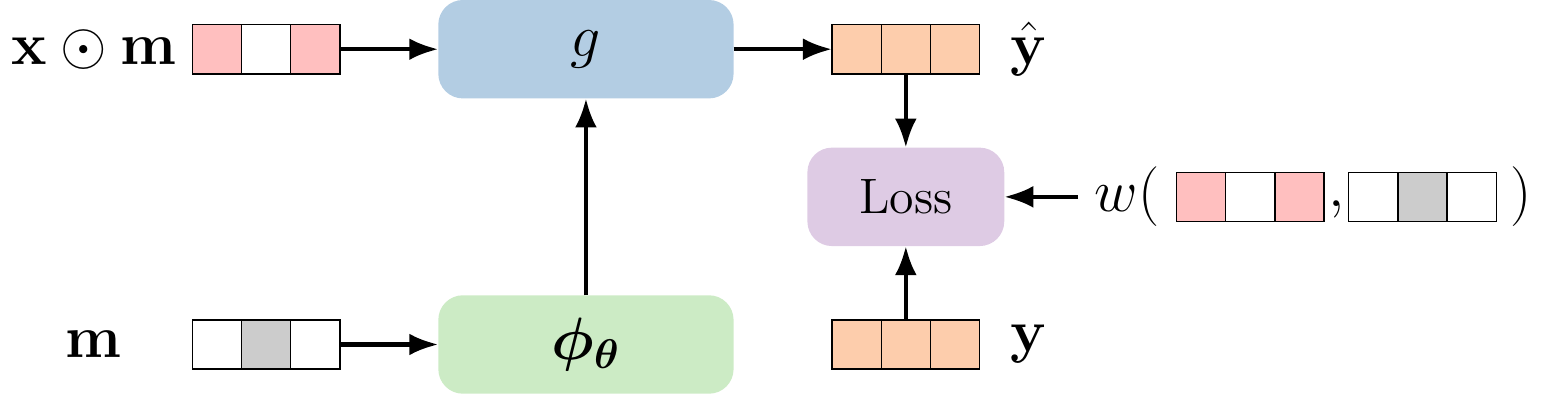}
	\caption{Prediction framework with sample reweighting.}
	\label{fig:framework}
\end{figure}

Note that the optimal predictor can be thought of as a function of the observed feature $\bx_\mathbf{m}$ parameterized by $\bphi$, where $\bphi$ is a function of $\mathbf{m}$, i.e.,  $g(\bx_\mathbf{m}, \mathbf{m}) = g_{\bphi(\mathbf{m})}(\bx_\mathbf{m})$. Learning all the $2^n$ optimal predictors then corresponds to learning the $2^n$ different values of $\bphi(\mathbf{m})$. The function $\bphi(\mathbf{m})$ corresponding to the optimal $g$ is very complicated in general. To address the exponential explosion problem, we adopt the common technique of approximating it by a simpler function $\bphi_{\btheta}(\mathbf{m})$ parameterized by $\btheta$, which can be implemented by a neural network  with parameter $\btheta$.
Since we will also implement $g$ by a neural network, we use  the zero-imputed feature  instead of $\bx_\mathbf{m}$ to uniformize the input size. The final form of our predictor is thus
\[
g(\bx_\mathbf{m},\mathbf{m})=g_{\bphi_{\btheta}(\mathbf{m})}(\bx\odot\mathbf{m}),
\]
where $\odot$ is element-wise multiplication and $\btheta$ is the parameter we need to learn. The framework is shown in Figure \ref{fig:framework}.

We illustrate the above framework by a simple example. Suppose $Y=\sum_{i=1}^n\alpha_iX_i$, where $X_i$'s are mutually independent except that $X_1=X_2$.  The optimal predictor is
\[
\E[Y\mid\bx_\mathbf{m},\mathbf{m}]=\phi_0+\sum_{i=1}^n \phi_i \cdot (\bx\odot\mathbf{m})_i,
\]
where 
\begin{equation}\label{eq:example}
\begin{aligned}
\phi_0 &=\bar m_1 \bar m_2(\alpha_1\E X_1+\alpha_2\E X_2)+\sum_{i=3}^n \bar m_i\alpha_i\E X_i,\\
\phi_1 &=(\alpha_1+\alpha_2\bar m_2)m_1,\quad
\phi_2 =(\alpha_2+\alpha_1\bar m_1)m_2,\\
\phi_i &=\alpha_im_i, \quad  \text{for}\; i=3,\dots,n.
\end{aligned}
\end{equation}
Note that, if $x_i$ is missing, the value of $\phi_i$ actually does not matter, as it is multiplied by $(\bx\odot \mathbf{m})_i = x_im_i=0$. Here the function $g$ is linear in $\bx\odot \mathbf{m}$ with the parameter $\bphi(\mathbf{m}) = (\phi_0(\mathbf{m}), \dots, \phi_n(\mathbf{m}))$ being quadratic in $\mathbf{m}$. We can easily parameterize $\bphi(\mathbf{m})$ by $\btheta\in\R^{(n+1)\times (n+1)\times (n+1)}$, where  $\phi_k(\mathbf{m})=\sum_{i,j}\theta_{ijk}m_im_j$. The corresponding $g$ is then 
\begin{equation}\label{eq:g}
g_{\bphi_{\btheta}(\mathbf{m})}(\bx\odot \mathbf{m})=\sum_{i,j,k}\theta_{ijk}x_km_im_jm_k,
\end{equation}
where $x_0 = m_0 = 1$. Here  approximation is not necessary as the optimal $\bphi(\mathbf{m})$ is simple enough. In general, however, $\bphi(\mathbf{m})$ may be very complex and approximation is necessary. For example, when the $X_i$'s are not independent, the $\bphi(\mathbf{m)}$ for the linear model can be an $n$-th order polynomial in $\mathbf{m}$ with $2^n$ parameters.

Note that the architecture of NeuMiss \citep{morvan2020neumiss} fits into our framework, although it does not explicitly consider mask distribution shift. NeuMiss uses a specific $g$ and $\bphi_{\btheta}$ that are carefully designed to approximate the analytical form of the optimal predictor for a linear label generation process with Gaussian feature. Our framework is more flexible and applies to more general settings, as shown in Section \ref{sec:experiment}.

\subsection{Decorrelation}\label{sec:decorrelation}

The use of $\bphi_{\btheta}(\mathbf{m})$ solves the exponential explosion problem, but it introduces another problem. As noted in prior work \citep{shen2018causally,kuang2018stable,kuang2020stable,shen2020stable,zhang2021deep,xu2022a}, the learned value of $\btheta$ can be affected by the correlation between the variables that we condition on in the optimal predictor, i.e., $\bX$ and $\bM$ in our case. This defies our original goal, as a learned value of $\btheta$ that performs well on the training distribution may not do so on the testing distribution, where the correlation changes. In what follows, we explain this problem in details.

We assume $\E[\bY\mid\bx_\mathbf{m},\mathbf{m}]$ is realizable, i.e., there exists a parameter $\btheta^*$ corresponding to the parameter of optimal predictor $\bphi$. In the above example, this assumption is satisfied by the form of $g$ in Equation \eqref{eq:g}, and $\btheta^*$ corresponds to the function $\bphi$ as given by Equation \eqref{eq:example}. In practice, we use neural networks for $g$ to approximately satisfy this assumption, which shows good empirical results in Section \ref{sec:experiment}.

Consider the population loss. $\btheta^*$ minimizes training loss
\begin{equation*}
    \ell_{tr}(\btheta)=\E_{(\bX,\bM,\bY)\sim p^{tr}}\|\bY-g_{\bphi_{\btheta}(\bM)}(\bX_\bM,\bM)\|_2^2\text{.}
\end{equation*}
Since parameter $\btheta^*$ corresponds to the parameter of optimal predictor, it also minimizes the above loss under testing distribution $p^{te}$, i.e., $\btheta^*$ generalizes under mask distribution shift.

Suppose $\hat\btheta$ is the parameter learned from training distribution, i.e. $\ell_{tr}(\hat \btheta)=\ell_{tr}(\btheta^*)$. $\hat\btheta$ can be different from $\btheta^*$, but they perform equally well on the training distribution. In fact, 
\begin{equation}\label{eq:equality}
g_{\bphi_{\hat \btheta}(\bM)}(\bX_\bM,\bM) = g_{\bphi_{\btheta^*}(\bM)}(\bX_\bM,\bM), \quad p^{tr}\text{-a.s.}\text{.}
\end{equation}

However, $\hat\btheta$ and $\btheta^*$ may not perform equally well on testing distribution $p^{tr}$, as formally stated in Theorem \ref{thm:unequal_testing_performance}, the proof of which is given in Appendix \ref{app:proof_unequal_testing_performance}.

\begin{theorem}\label{thm:unequal_testing_performance}
 In the realizable case, there exists a problem instance, in which there exists a $\hat\btheta$ other than $\btheta^*$ such that Equation \eqref{eq:equality} holds $p^{tr}\text{-a.s.}$ but not $p^{te}\text{-a.s.}$.
\end{theorem}

Such $\hat\btheta$ is undesirable, since it cannot generalize as well as $\btheta^*$. To address this problem, decorrelation among $\bX$ and $\bM$ may help. We explain the reason with an example below.

Consider the example in Section \ref{sec:prediction_framework}. If the entries of $\bX$ and the entries of $\bM$ are mutually independent, the only solution for $\hat\btheta$ is the $\btheta^*$ corresponding to Equation \eqref{eq:example}. The proof is given in Appendix \ref{app:proof_decorrelation}.
Thus we can perform decorrelation to help obtain the desired parameter $\btheta^*$.

Under MCAR, we decorrelate the entries of $\bX$ and those of $\bM$ respectively; under MAR, we further decorrelate between $\bX$ and $\bM$. In this paper, we follow \citet{zhang2021deep} to decorrelate by sample reweighting. Specifically, we learn a weighting function $w(\bx_\mathbf{m},\mathbf{m})$ by minimizing the correlation under distribution $\tilde{p}_w$, where
\begin{equation*}
    \tilde{p}_w(\bx,\mathbf{m},y)=w(\bx_\mathbf{m},\mathbf{m})p^{tr}(\bx,\mathbf{m},y)\text{.}
\end{equation*}
The learned weighting function will be normalized to make $\tilde{p}_w$ well-defined. Under $\tilde{p}_w$, the conditional label distribution $p(\by\mid\bx_{\mathbf{m}},\mathbf{m})$ does not change. Inspired by \citet{zhang2021deep}, we measure the correlation empirically by partial cross-covariance matrix with Random Fourier Feature \citep{strobl2019approximate}. We adapt it to the case of incomplete data as follows.

The partial cross-covariance matrix of $X_k$ and $X_l$, denoted by $\Sigma_{X_k,X_l;w}^\prime$, is computed with only samples in which $X_k$ and $X_l$ are both observed:
\begin{equation*}
    \small
    \begin{aligned}
        \Sigma_{X_k,X_l;w}^\prime=\frac{1}{N^{kl}-1}\sum_{i=1}^{N^{kl}}\biggl[ & \biggl(w_i\bu(X_k^{(i)})-\frac{1}{N^{k}}\sum_{j=1}^{N^{k}}w_j\bu(X_k^{(j)})\biggr)^\top \\
          & \biggl(w_i\bv(X_l^{(i)})-\frac{1}{N^{l}}\sum_{j=1}^{N^{l}}w_j\bv(X_l^{(j)})\biggr)\biggr]\text{,}
    \end{aligned}
\end{equation*}
where $N^{kl}$ is the number of samples in which $X_k$ and $X_l$ are both observed, $N^k$ and $N^l$ are the number of samples with observed $X_k$ and $X_l$, respectively, $w_i=w(\bx_{\mathbf{m}^{(i)}}^{(i)},\mathbf{m}^{(i)})$, and $\bu$, $\bv$ are function vectors with elements drawn from the space of Random Fourier Feature \citep{strobl2019approximate}, the definition of which is given in Appendix \ref{app:rff}. Similarly, $\Sigma_{X_k,M_l;w}^\prime$ is computed with only samples in which $X_k$ is observed, and $\Sigma_{M_k,M_l;w}^\prime$ is computed with all the $N$ samples. See Appendix \ref{app:rff} for their detailed forms.

When decorrelating the entries of $\bX$ and those of $\bM$, we optimize sample weight $\bw\in\R_+^N$ by minimizing the correlation of all pairs of $X_i$ and $X_j$ and all pairs of $M_i$ and $M_j$:
\begin{equation*}
    \begin{aligned}
        \min_{\bw\in\R^+} & \sum_{1\leq k<l\leq n}\|\Sigma_{X_k,X_l;w}^\prime\|_F^2+\sum_{1\leq k<l\leq n}\|\Sigma_{M_k,M_l;w}^\prime\|_F^2 \\
          & +\gamma\frac{\Std(\bw)}{1/N\sum_{i=1}^N w_i}\text{,}
    \end{aligned}
\end{equation*}
where $\Std(\bw)$ is standard deviation of the $w_i$'s. The last regularization term is used to prevent assigning very large weights to a small proportion of samples. Hyper-parameter $\gamma$ is coefficient for the regularization term. Under MAR, we further add $\sum_{1\leq k,l\leq n}\|\Sigma_{X_k,M_l;w}^\prime\|_F^2$ into the objective to decorrelate between $\bX$ and $\bM$.

With the learned weighting function, we can conduct regression under the weighted distribution $\tilde{p}_w$:
\begin{equation*}
	\min_{\btheta}\E_{\tilde{p}_w}[(Y-g_{\bphi_{\btheta}(\bM)}(\bX_\bM,\bM))^2]\text{.}
\end{equation*}
In the ideal case, the entries of $\bX$ and those of $\bM$ are mutually independent under $\tilde{p}_w$. Then the learned $\btheta$ will not be affected by the correlation and can generalize under mask distribution shift. In practice, we can equivalently conduct weighted regression under the training distribution:
\begin{equation*}
	\min_{\btheta}\E_{p^{tr}}[w(\bX_\bM,\bM)(Y-g_{\bphi_{\btheta}(\bM)}(\bX_\bM,\bM))^2]\text{.}
\end{equation*}
How decorrelation is combined with the prediction framework is shown in Figure \ref{fig:framework}.


\section{Experiment}\label{sec:experiment}

\begin{table*}[t]
    \centering
    \caption{Performance on Gaussian-Mix feature with MAR mask when trained under 50\% missing level. The values for Optimal are the RMSE, while the other values are the gap between the RMSE of the corresponding method and that of  Optimal with the same experimental setup. \textbf{Bold} and \underline{underline} represent the best and second best along each column, respectively. Superscript $^*$ indicates when training and testing missing levels are the same. These marks will also be used in the other tables.}
    \label{tbl:gaussian-mix_mar_fixed-training}
    \begin{tabular}{@{}clrrrrrrrrr}
        \toprule
        \multicolumn{2}{c}{\multirow{2.5}*{\textbf{Method}}} & \multicolumn{9}{c}{\textbf{Testing Missing Level}} \\ \cmidrule{3-11}
        ~ & ~ & \textbf{10\%} & \textbf{20\%} & \textbf{30\%} & \textbf{40\%} & \textbf{50\%} & \textbf{60\%} & \textbf{70\%} & \textbf{80\%} & \textbf{90\%} \\
        \midrule
        \multirow{8}*{\specialcell{\textbf{Gap to}\\\textbf{Optimal}}} & Partial VAE & 1541.52 & 1244.90 & 1172.68 & 961.18 & 793.32$^*$ & 898.04 & 1075.69 & 1087.19 & 1112.42 \\
        ~ & MIWAE & 1331.95 & 1029.18 & 1047.14 & 910.77 & 743.63$^*$ & 841.89 & 820.39 & 844.80 & 844.17 \\
        ~ & P-BiGAN & 1382.71 & 1123.64 & 1041.63 & 923.53 & 713.85$^*$ & 856.39 & 845.39 & 867.93 & 979.45 \\
        ~ & NeuMiss & \underline{714.19} & \underline{520.26} & \underline{523.10} & \underline{432.99} & \underline{293.14}$^*$ & \underline{574.80} & \underline{601.74} & \underline{646.53} & \underline{679.99} \\
        \cmidrule{2-11}
        ~ & DWR & 1485.42 & 1170.15 & 1142.32 & 1033.32 & 893.64$^*$ & 962.61 & 969.56 & 966.85 & 963.52 \\
        ~ & SRDO & 1385.00 & 1116.82 & 1080.93 & 988.86 & 843.85$^*$ & 919.26 & 899.21 & 903.82 & 866.09 \\
        ~ & StableNet & 1256.32 & 967.22 & 989.67 & 878.97 & 743.25$^*$ & 818.16 & 807.50 & 750.89 & 788.56 \\
        \cmidrule{2-11}
        ~ & {\ours} & \textbf{431.16} & \textbf{342.70} & \textbf{309.09} & \textbf{278.97} & \textbf{282.75}$^*$ & \textbf{319.58} & \textbf{405.60} & \textbf{412.03} & \textbf{467.98} \\
        \midrule
        \multicolumn{2}{c}{\textbf{Optimal}} & 904.87 & 1096.36 & 1108.73 & 1224.50 & 1324.00 & 1375.16 & 1518.59 & 1626.74 & 1705.54 \\
        \bottomrule
    \end{tabular}
\end{table*}

\subsection{Datasets}\label{sec:datasets}

We evaluate {\ours} on synthetic and real-world datasets.

\textbf{Gaussian}. Following \citet{morvan2021whats}, we generate feature $\bX$ from multivariate Gaussian distribution. The mean values are drawn from standard Gaussian distribution, and the covariance matrix is generated by $\Sigma=\bB\bB^\top+\bD$, where the entries of $\bB\in\R^{n\times 0.7n}$ are drawn from standard Gaussian distribution and $\bD\in\R^{n\times n}$ is diagonal with values uniformly drawn from $[10^{-2},10^{-1}]$. The feature dimension $n=50$.

\textbf{Gaussian-Ind}. The feature generation is the same as Gaussian, except that we make the entries of $\bX$ mutually independent by using a diagonal covariance matrix.

\textbf{Gaussian-Mix}. We generate feature $\bX$ from the more general Gaussian mixture model with $3$ components, each generated in the same way as Gaussian. The proportion of the $i$-th component $\pi_i$ is uniformly drawn from $[0, 1)$ and normalized by $\pi_i/\sum_{i=1}^3\pi_i$.

For the above 3 synthetic features, the label generation process is linear: $Y=\alpha_0+\sum_{i=1}^n\alpha_i X_i+\epsilon$, where $\epsilon$ is a Gaussian noise such that the signal-to-noise ratio is 10. Different from \citet{morvan2020neumiss,morvan2021whats} where all $\alpha_i$'s except $\alpha_0$ are equal, we draw more general $\alpha_i$'s from Gaussian distribution.

\textbf{House Sales}. Following \citet{shen2020stable}, we use dataset of house sales in King County, USA, which contains $n=16$ features and a scalar house price as label.

\textbf{MNIST} \citep{lecun1998gradient-based}. The MNIST dataset of handwritten digit images. Given incomplete image, we aim to predict the complete image.

\textbf{Traffic} \citep{didi2018city}. Average traffic speed within every hour from 1343 roads in the city of Chengdu, China, in 2018. We build a graph for the dataset, where nodes represent the roads and edges indicate the adjacency of roads. Note that this dataset is naturally incomplete. Given incomplete history, we aim to predict the future traffic speed.

All the datasets except the Traffic dataset are complete. We generate incomplete datasets by imposing mask on the complete samples according to the missing patterns in Section \ref{sec:missing_patterns}.

\begin{table*}[t]
    \centering
    \caption{Performance on Gaussian-Mix feature with MAR mask when tested under 50\% missing level.}
    \label{tbl:gaussian-mix_mar_fixed-testing}
    \begin{tabular}{@{}clrrrrrrrrr}
        \toprule
        \multicolumn{2}{c}{\multirow{2.5}*{\textbf{Method}}} & \multicolumn{9}{c}{\textbf{Training Missing Level}} \\ \cmidrule{3-11}
        ~ & ~ & \textbf{10\%} & \textbf{20\%} & \textbf{30\%} & \textbf{40\%} & \textbf{50\%} & \textbf{60\%} & \textbf{70\%} & \textbf{80\%} & \textbf{90\%} \\
        \midrule
        \multirow{8}*{\specialcell{\textbf{Gap to}\\\textbf{Optimal}}} & Partial VAE & 1441.03 & 946.28 & 951.77 & 880.44 & 793.32$^*$ & 1048.26 & 1581.51 & 1608.35 & 1897.15 \\
        ~ & MIWAE & 1092.03 & 806.06 & 816.18 & 753.13 & 743.63$^*$ & 883.19 & 1333.60 & 1235.24 & 1674.31 \\
        ~ & P-BiGAN & 1136.98 & 840.27 & 849.73 & 771.86 & 713.85$^*$ & 947.65 & 1215.25 & 1255.63 & 1709.70 \\
        ~ & NeuMiss & \underline{687.57} & \underline{650.05} & \underline{529.37} & \underline{424.46} & \underline{293.14}$^*$ & \underline{692.26} & \underline{744.01} & \underline{850.52} & \underline{1023.01} \\
        \cmidrule{2-11}
        ~ & DWR & 1111.11 & 957.66 & 962.37 & 904.77 & 893.64$^*$ & 1054.67 & 1312.80 & 1434.32 & 1902.80 \\
        ~ & SRDO & 991.91 & 939.84 & 890.83 & 884.84 & 843.85$^*$ & 1029.93 & 1362.18 & 1359.29 & 1897.69 \\
        ~ & StableNet & 856.08 & 847.25 & 866.97 & 823.89 & 743.25$^*$ & 889.24 & 1336.49 & 1410.46 & 1697.40 \\
        \cmidrule{2-11}
        ~ & {\ours} & \textbf{337.14} & \textbf{326.49} & \textbf{296.93} & \textbf{297.95} & \textbf{282.75}$^*$ & \textbf{421.15} & \textbf{480.80} & \textbf{584.96} & \textbf{716.63} \\
        \midrule
        \multicolumn{2}{c}{\textbf{Optimal}} & & & & & 1324.00 & & & & \\
        \bottomrule
    \end{tabular}
\end{table*}

\begin{table*}[t]
    \centering
    \caption{Performance on House Sales dataset with MAR mask when trained under 50\% missing level (unit: \$10000).}
    \label{tbl:house-sales_mar_fixed-training}
    \begin{tabular}{@{}clrrrrrrrrr}
        \toprule
        \multicolumn{2}{c}{\multirow{2.5}*{\textbf{Method}}} & \multicolumn{9}{c}{\textbf{Testing Missing Level}} \\ \cmidrule{3-11}
        ~ & ~ & \textbf{10\%} & \textbf{20\%} & \textbf{30\%} & \textbf{40\%} & \textbf{50\%} & \textbf{60\%} & \textbf{70\%} & \textbf{80\%} & \textbf{90\%} \\
        \midrule
        \multirow{8}*{\specialcell{\textbf{Gap to}\\\textbf{{\ours}-ID}}} & Partial VAE & 23.37 & 18.94 & 17.94 & 14.85 & 8.46$^*$ & 19.70 & 21.87 & 46.05 & 52.53 \\
        ~ & MIWAE & 27.14 & 23.92 & 22.90 & 21.16 & 13.73$^*$ & 21.52 & 25.73 & 49.95 & 52.56 \\
        ~ & P-BiGAN & 32.91 & 25.62 & 23.82 & 23.10 & 16.36$^*$ & 26.29 & 27.44 & 56.17 & 62.87 \\
        ~ & NeuMiss & 25.89 & 20.19 & 17.89 & \underline{14.48} & \underline{7.67}$^*$ & 17.67 & 21.26 & 46.55 & 53.06 \\
        \cmidrule{2-11}
        ~ & DWR & 38.36 & 26.80 & 25.22 & 23.70 & 11.09$^*$ & 26.23 & 27.62 & 61.13 & 71.49 \\
        ~ & SRDO & 29.46 & 22.49 & 18.89 & 15.16 & 6.88$^*$ & \underline{15.65} & 24.86 & 51.77 & 55.68 \\
        ~ & StableNet & \underline{21.41} & \underline{15.18} & \underline{14.78} & 14.53 & 8.46$^*$ & 18.55 & \underline{15.96} & \underline{38.09} & \underline{46.99} \\
        \cmidrule{2-11}
        ~ & {\ours} & \textbf{14.68} & \textbf{10.16} & \textbf{8.46} & \textbf{7.03} & \textbf{0.00}$^*$ & \textbf{9.51} & \textbf{10.31} & \textbf{21.90} & \textbf{34.90} \\
        \midrule
       	\multicolumn{2}{c}{\textbf{{\ours}-ID}} & 20.85 & 22.80 & 24.36 & 28.77 & 32.89 & 36.67 & 39.23 & 42.90 & 48.16 \\
        \bottomrule
    \end{tabular}
\end{table*}


\begin{table*}[t]
    \centering
    \caption{Performance on MNIST dataset with MAR mask when trained under 50\% missing level.}
    \label{tbl:mnist_mar_fixed-training}
    \begin{tabular}{@{}clrrrrrrrrr}
        \toprule
        \multicolumn{2}{c}{\multirow{2.5}*{\textbf{Method}}} & \multicolumn{9}{c}{\textbf{Testing Missing Level}} \\ \cmidrule{3-11}
        ~ & ~ & \textbf{10\%} & \textbf{20\%} & \textbf{30\%} & \textbf{40\%} & \textbf{50\%} & \textbf{60\%} & \textbf{70\%} & \textbf{80\%} & \textbf{90\%} \\
        \midrule
        \multirow{8}*{\specialcell{\textbf{Gap to}\\\textbf{{\ours}-ID}}} & Partial VAE & 21.61 & 17.51 & 16.58 & 13.74 & 7.82$^*$ & 18.22 & 20.22 & 42.58 & 48.57 \\
        ~ & MIWAE & 25.09 & 22.12 & 21.18 & 19.57 & 12.69$^*$ & 19.90 & 23.79 & 46.18 & 48.60 \\
        ~ & P-BiGAN & 30.43 & 23.69 & 22.03 & 21.36 & 15.13$^*$ & 24.31 & 25.38 & 51.94 & 58.13 \\
        ~ & NeuMiss & 23.94 & 18.67 & 16.54 & \underline{13.39} & \underline{7.09}$^*$ & 16.34 & 19.66 & 43.04 & 49.06 \\
        \cmidrule{2-11}
        ~ & DWR & 35.47 & 24.78 & 23.32 & 21.92 & 10.26$^*$ & 24.25 & 25.54 & 56.52 & 66.10 \\
        ~ & SRDO & 27.24 & 20.79 & 17.47 & 14.01 & 6.36$^*$ & \underline{14.47} & 22.99 & 47.87 & 51.49 \\
        ~ & StableNet & \underline{19.79} & \underline{14.03} & \underline{13.67} & 13.43 & 7.82$^*$ & 17.15 & \underline{14.76} & \underline{35.22} & \underline{43.45} \\
        \cmidrule{2-11}
        ~ & {\ours} & \textbf{12.33} & \textbf{8.75} & \textbf{7.05} & \textbf{6.65} & \textbf{0.00}$^*$ & \textbf{8.25} & \textbf{9.99} & \textbf{18.73} & \textbf{27.47} \\
        \midrule
       	\multicolumn{2}{c}{\textbf{{\ours}-ID}} & 19.27 & 22.14 & 24.08 & 27.96 & 31.93 & 34.95 & 38.06 & 41.65 & 46.76 \\
        \bottomrule
    \end{tabular}
\end{table*}


\begin{table*}[t]
    \centering
    \caption{Performance on Traffic dataset. All the values are exact RMSE (unit: km/h).}
    \label{tbl:traffic}
    \begin{tabular}{ccccccccc}
        \toprule
        Partial VAE & MIWAE & P-BiGAN & NeuMiss & DWR & SRDO & StableNet & {\ours} & {\ours}-ID \\
        \midrule
       	16.08 & 14.82 & 15.79 & 15.10 & 15.93 & 15.72 & 13.97 & \textbf{11.84} & 8.39 \\
        \bottomrule
    \end{tabular}
\end{table*}

\subsection{Missing Patterns}\label{sec:missing_patterns}

We design the following 3 missing patterns. There are 9 missing levels, denoted by $r$, from 10\% to 90\% at a step of 10\%. Missing patterns examples are given in Appendix \ref{app:missing_pattern_example}.

\textbf{MCAR-Ind}. We generate mask $\bM$ that is independent of feature $\bX$, and the entries of mask are mutually independent. For each sample, its sample missing rate $r_s$ has 80\% to be $r$ and 2.5\% to be one of the other 8 levels respectively. Each entry is independently missing with probability $r_s$.

\textbf{MCAR}. We generate mask $\bM$ that is independent of feature $\bX$, but the entries of mask can be dependent. The sample missing rate $r_s$ is determined in the same way as MCAR-Ind. In each sample, following \citet{li2020learning}, we generate a window of length $\lfloor n\cdot r_s\rfloor$ at a random position, where the $\lfloor n\cdot r_s\rfloor$ consecutive features in the window are missing.

\textbf{MAR}. Following \citet{morvan2020neumiss,morvan2021whats}, we generate feature-dependent mask, and the entries of mask can also be dependent. First, randomly selected 10\% features are set to be observed in all the samples. The mask on the other features are generated according to a model whose parameters depend on the selected features. The sample missing rate $r_s$ means the missing proportion of the other 90\% features, which is determined in the same way as MCAR-Ind.

\subsection{Baselines}\label{sec:baselines}

We compare {\ours} with two categories of baselines. (1) state-of-the-art methods on prediction with incomplete data: NeuMiss \citep{morvan2020neumiss,morvan2021whats}, Partial VAE \citep{ma2019eddi}, MIWAE \citep{mattei2019miwae} and P-BiGAN \citep{li2020learning}; (2) state-of-the-art methods on generalization under agnostic feature distribution shift: DWR \citep{kuang2020stable}, SRDO \citep{shen2020stable} and StableNet \citep{zhang2021deep}. For Partial VAE, MIWAE and P-BiGAN, we treat the label as missing feature. For DWR, SRDO and StableNet, we use mean-imputed feature as input. See Appendix \ref{app:implementation_details} for the implementation details of these methods.

We use the commonly adopted Root Mean Square Error (RMSE) as evaluation metric throughout.

\subsection{Performance on Synthetic Dataset}\label{sec:synthetic_performance}

We compare {\ours} with baselines under different missing patterns and missing levels. The models are trained under a specific missing level and tested under all the missing levels. The difference in missing level represents the mask distribution shift. Due to space limit, we only present the results of missing rate shift on Gaussian-Mix feature with MAR mask; see Appendix \ref{app:other-performance} and \ref{app:pattern_shift} for the missing rate shift on other settings and missing pattern shift. We also present the performance of optimal predictor $\E[Y\mid\bx_\mathbf{m},\mathbf{m}]$. It can be derived from known feature distribution and label generation process, without which the performance of optimal predictor is not reachable.

We will show the results from two views: results with fixed training missing level that reflects the generalizability of methods, and results with fixed testing missing level that reflects the robustness to the mask distribution of training data.

\textbf{Fixing Training Missing Level}. We show the results when trained under 50\% missing level and tested under all the missing levels in Table \ref{tbl:gaussian-mix_mar_fixed-training}; the other settings are similar. Since the amount of observed data is different between missing levels, which influences the prediction error, we use the gap to optimal to reflect the generalization performance.

Error of the optimal predictor increases with missing level, since there are less observed data under higher missing level. When the mask distribution shifts, {\ours} has the best generalization performance, reducing the gap to optimal of the second best, NeuMiss in this case,  by 31\%-44\%.
When there is no mask distribution shift, {\ours} still has the best performance in this case, slightly outperforming NeuMiss by 4\% in terms of the gap to optimal. As shown in Appendix \ref{app:other-performance},  for Gaussian features with MCAR or MAR mask, NeuMiss can slightly outperform {\ours}, by at most 2\%, in the absence of mask distribution shift. This is not too surprising, as these  are the cases that NeuMiss is tailored for.

\textbf{Fixing Testing Missing Level}. We show the results when tested under 50\% missing level and trained under all the missing levels in Table \ref{tbl:gaussian-mix_mar_fixed-testing}; the other settings are similar. When the mask distribution shifts, {\ours} achieves the best performance, reducing the gap to optimal of the second best NeuMiss by 30\%-50\%.
Reduction is especially large when trained under 10\% missing level, since the optimal predictors are more learnable under lower missing level. The results show that {\ours} is robust to the quality, i.e., mask distribution, of training data and can generalize from various training mask distributions.

\subsection{Performance on Real-World Dataset}\label{sec:real-world_performance}

\textbf{House Sales}. We show the results on House Sales dataset when trained under 50\% missing level in Table \ref{tbl:house-sales_mar_fixed-training}. Without prior knowledge about label generation process, the optimal predictor cannot be derived. {\ours} has the best performance when trained in distribution, i.e., trained under testing distribution. Instead of Optimal, we show our in-distribution performance, which is denoted by `{\ours}-ID'.

Error of {\ours}-ID also increases with missing level. When the mask distribution shifts, {\ours} achieves the best performance, reducing the gap to {\ours}-ID of the second best, usually StableNet, by 26\%-50\%, with an average of 38\%, which demonstrates our generalizability on real-world data.

\textbf{MNIST}. We show the results on MNIST dataset when trained under 50\% missing level in Table \ref{tbl:mnist_mar_fixed-training}. When the mask distribution shifts, {\ours} achieves the best performance, reducing the gap to {\ours}-ID of the second best, usually StableNet, by 28\%-50\%, with an average of 41\%.

\textbf{Traffic}. Table \ref{tbl:traffic} shows the results on Traffic dataset, which is naturally incomplete. The training and testing missing rates are 37\% and 23\%. Since ground truth of missing entries is unavailable, RMSE is only computed on observed entries. The value for {\ours}-ID is the training error, which only serves as a reference for our comparison. Note that {\ours} achieves the best performance, reducing the gap to {\ours}-ID of the second best, StableNet,  by 38\%, which shows that {\ours} can be applied to real-world datasets with complex missingness.

\subsection{Ablation Study}\label{sec:ablation_study}

\begin{figure}[t]
	\centering
	\includegraphics[width=0.7\columnwidth]{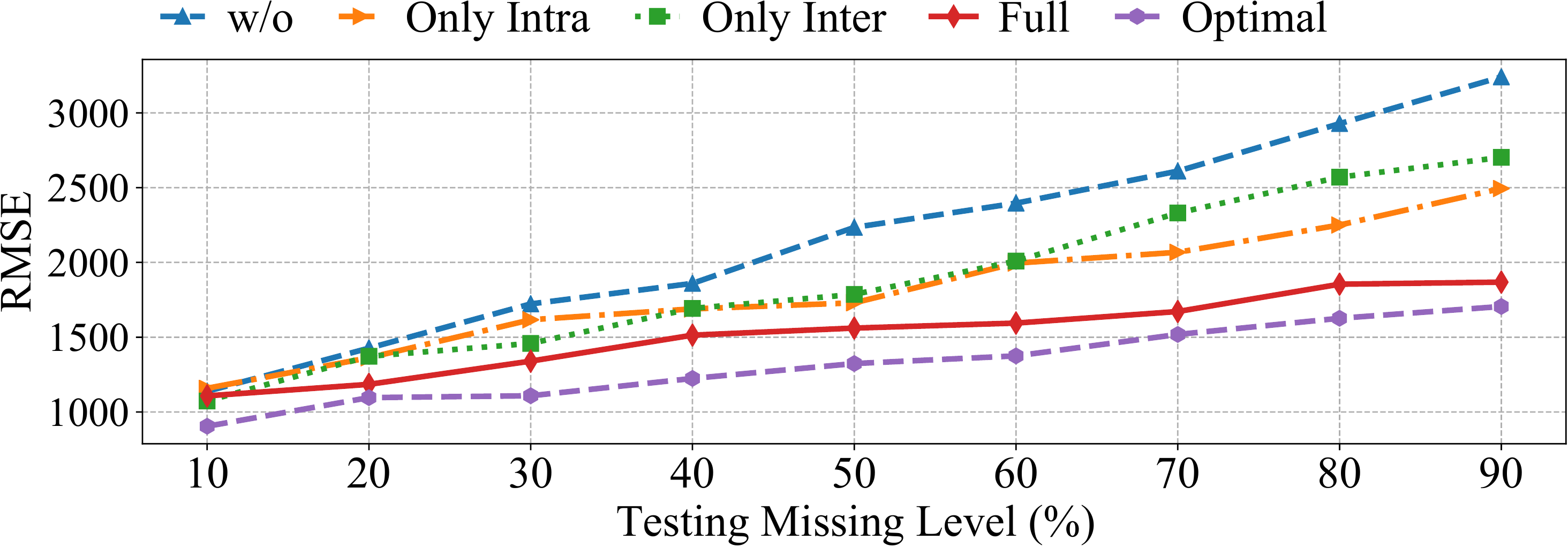}
	\caption{Comparison with ablated variants: only decorrelating the entries of $\bX$ and $\bM$ respectively (only intra), only decorrelating between $\bX$ and $\bM$ (only inter), and no decorrelaion (w/o). The same notations are used in Figure \ref{fig:unneeded_decorrelation}.}
	\label{fig:ablation}
\end{figure}

We study the efficacy of decorrelation by comparing with ablated variants that only decorrelates the entries of $\bX$ and $\bM$ respectively (only intra), only decorrelates between $\bX$ and $\bM$ (only inter) and does not decorrelate (w/o). The result on Gaussian-Mix feature with MAR mask when trained under 10\% missing level is given in Figure \ref{fig:ablation}. Our method with full decorrelation achieves the best performance especially when mask distribution shift becomes stronger. The gap of the second best to optimal is reduced by 55\% on average and by 83\% under 90\% missing level.

\begin{figure}[t]
	\centering
	\includegraphics[width=0.7\columnwidth]{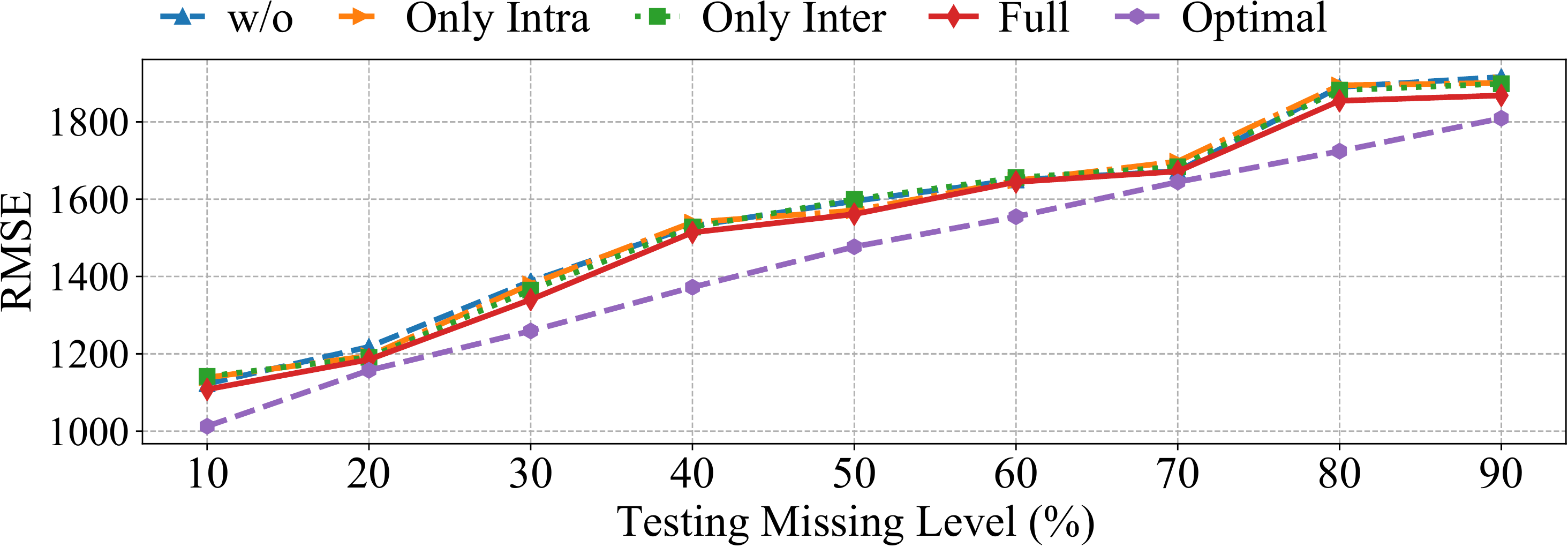}
	\caption{Influence of unneeded decorrelation.}
	\vspace{-3ex}
	\label{fig:unneeded_decorrelation}
\end{figure}

We also study the influence of decorrelation when it is not needed. The result on Gaussian-Ind feature with MCAR-Ind mask when trained under 10\% missing level is given in Figure \ref{fig:unneeded_decorrelation}. The performance with unneeded decorrelation is close to that without decorrelation. For real-world data with unknown correlation, we can always conduct decorrelation.


\section{Conclusion}\label{sec:conclusion}

In this paper, we propose a novel method {\ours} for the problem of \textit{prediction with incomplete data under agnostic mask distribution shift}. We leverage the observation that for each mask, there is an invariant optimal predictor. We approximate the optimal predictors jointly using a double parameterization technique. We also perform decorrelation to minimize the side effect caused by the intra-mask correlation and the correlation between features and mask. Extensive experiments are conducted on both synthetic and real-world datasets. The results show that {\ours} is robust and outperforms state-of-the-art methods under agnostic mask distribution shift.

\newpage

\section*{Acknowledgments}

This work was supported by NSF China (No. 62072302, U20A20181, U21A20519, 42050105, 62262018) and the Open Research Project of the State Key Laboratory of Media Convergence and Communication, Communication University of China, China (No.SKLMCC2021KF011).

\bibliographystyle{named}
\bibliography{contents/bib}

\begin{thebibliography}{}

\bibitem[\protect\citeauthoryear{Burda \bgroup \em et al.\egroup
  }{2016}]{burda2016importance}
Yuri Burda, Roger~B. Grosse, and Ruslan Salakhutdinov.
\newblock Importance weighted autoencoders.
\newblock In {\em The 4th International Conference on Learning
  Representations}, San Juan, Puerto Rico, 2016.

\bibitem[\protect\citeauthoryear{DiDiChuxing}{2018}]{didi2018city}
DiDiChuxing.
\newblock City traffic index.
\newblock https://gaia.didichuxing.com/, 2018.

\bibitem[\protect\citeauthoryear{Donahue \bgroup \em et al.\egroup
  }{2017}]{donahue2017adversarial}
Jeff Donahue, Philipp Krahenbuhl, and Trevor Darrell.
\newblock Adversarial feature learning.
\newblock In {\em The 5th International Conference on Learning
  Representations}, Toulon, France, 2017.

\bibitem[\protect\citeauthoryear{Goodfellow \bgroup \em et al.\egroup
  }{2014}]{goodfellow2014generative}
Ian Goodfellow, Jean Pouget-Abadie, Mehdi Mirza, Bing Xu, David Warde-Farley,
  Sherjil Ozair, Aaron Courville, and Yoshua Bengio.
\newblock Generative adversarial nets.
\newblock In {\em Advances in Neural Information Processing Systems}, pages
  2672--2680, Montreal, Canada, December 2014.

\bibitem[\protect\citeauthoryear{Kingma and
  Welling}{2014}]{kingma2014auto-encoding}
Diederik~P. Kingma and Max Welling.
\newblock Auto-encoding variational bayes.
\newblock In {\em The 2nd International Conference on Learning
  Representations}, Banff, Canada, 2014.

\bibitem[\protect\citeauthoryear{Kuang \bgroup \em et al.\egroup
  }{2018}]{kuang2018stable}
Kun Kuang, Peng Cui, Susan Athey, Ruoxuan Xiong, and Bo~Li.
\newblock Stable prediction across unknown environments.
\newblock In {\em Proceedings of the 24th ACM SIGKDD International Conference
  on Knowledge Discovery \& Data Mining}, pages 1617--1626, London, UK, 2018.

\bibitem[\protect\citeauthoryear{Kuang \bgroup \em et al.\egroup
  }{2020}]{kuang2020stable}
Kun Kuang, Ruoxuan Xiong, Peng Cui, Susan Athey, and Bo~Li.
\newblock Stable prediction with model misspecification and agnostic
  distribution shift.
\newblock In {\em The 34th AAAI Conference on Artificial Intelligence}, pages
  4485--4492, New York, USA, 2020.

\bibitem[\protect\citeauthoryear{Lecun \bgroup \em et al.\egroup
  }{1998}]{lecun1998gradient-based}
Yann Lecun, Leon Bottou, Yoshua Bengio, and Patrick Haffner.
\newblock Gradient-based learning applied to document recognition.
\newblock {\em Proceedings of the IEEE}, 86(11):2278--2324, 1998.

\bibitem[\protect\citeauthoryear{Li and Marlin}{2020}]{li2020learning}
Steven Cheng-Xian Li and Benjamin Marlin.
\newblock Learning from irregularly-sampled time series: A missing data
  perspective.
\newblock In {\em Proceedings of the 37th International Conference on Machine
  Learning}, pages 5937--5946, Virtual Event, 2020.

\bibitem[\protect\citeauthoryear{Li \bgroup \em et al.\egroup
  }{2019}]{li2019learning}
Steven Cheng-Xian Li, Bo~Jiang, and Benjamin Marlin.
\newblock Learning from incomplete data with generative adversarial networks.
\newblock In {\em Proceedings of the 7th International Conference on Learning
  Representations}, New Orleans, USA, May 2019.

\bibitem[\protect\citeauthoryear{Little and
  Rubin}{1986}]{little1986statistical}
Roderick J.~A. Little and Donald~B. Rubin.
\newblock {\em Statistical Analysis with Missing Data}.
\newblock John Wiley \& Sons, Inc., USA, 1986.

\bibitem[\protect\citeauthoryear{Ma \bgroup \em et al.\egroup
  }{2019}]{ma2019eddi}
Chao Ma, Sebastian Tschiatschek, Konstantina Palla, Jose~Miguel
  Hernandez-Lobato, Sebastian Nowozin, and Cheng Zhang.
\newblock {EDDI}: Efficient dynamic discovery of high-value information with
  partial vae.
\newblock In {\em Proceedings of the 36th International Conference on Machine
  Learning}, pages 4234--4243, Long Beach, USA, 2019.

\bibitem[\protect\citeauthoryear{Mattei and Frellsen}{2019}]{mattei2019miwae}
Pierre-Alexandre Mattei and Jes Frellsen.
\newblock {MIWAE}: Deep generative modelling and imputation of incomplete data
  sets.
\newblock In {\em Proceedings of the 36th International Conference on Machine
  Learning}, pages 4413--4423, Long Beach, USA, 2019.

\bibitem[\protect\citeauthoryear{Morvan \bgroup \em et al.\egroup
  }{2020}]{morvan2020neumiss}
Marine~Le Morvan, Julie Josse, Thomas Moreau, Erwan Scornet, and Gael
  Varoquaux.
\newblock {NeuMiss} networks: Differentiable programming for supervised
  learning with missing values.
\newblock In {\em Proceedings of the 34th International Conference on Neural
  Information Processing Systems}, page 5980–5990, Vancouver, Canada, 2020.

\bibitem[\protect\citeauthoryear{Morvan \bgroup \em et al.\egroup
  }{2021}]{morvan2021whats}
Marine~Le Morvan, Julie Josse, Erwan Scornet, and Gael Varoquaux.
\newblock What's a good imputation to predict with mmissing values?
\newblock In {\em Proceedings of the 35th International Conference on Neural
  Information Processing Systems}, pages 11530--11540, Virtual Event, 2021.

\bibitem[\protect\citeauthoryear{Pan and Yang}{2010}]{pan2010a}
Sinno~Jialin Pan and Qiang Yang.
\newblock A survey on transfer learning.
\newblock {\em IEEE Transactions on Knowledge and Data Engineering},
  22(10):1345--1359, 2010.

\bibitem[\protect\citeauthoryear{Shen \bgroup \em et al.\egroup
  }{2018}]{shen2018causally}
Zheyan Shen, Peng Cui, Kun Kuang, Bo~Li, and Peixuan Chen.
\newblock Causally regularized learning with agnostic data selection bias.
\newblock In {\em Proceedings of the 26th ACM International Conference on
  Multimedia}, page 411–419, Seoul, Republic of Korea, 2018.

\bibitem[\protect\citeauthoryear{Shen \bgroup \em et al.\egroup
  }{2020}]{shen2020stable}
Zheyan Shen, Peng Cui, Tong Zhang, and Kun Kuang.
\newblock Stable learning via sample reweighting.
\newblock In {\em The 34th AAAI Conference on Artificial Intelligence}, pages
  5692--5699, New York, USA, 2020.

\bibitem[\protect\citeauthoryear{Strobl \bgroup \em et al.\egroup
  }{2019}]{strobl2019approximate}
Eric~V. Strobl, Kun Zhang, and Shyam Visweswaran.
\newblock Approximate kernel-based conditional independence tests for fast
  non-parametric causal discovery.
\newblock {\em Journal of Causal Inference}, 7(1), 2019.

\bibitem[\protect\citeauthoryear{Sugiyama \bgroup \em et al.\egroup
  }{2012}]{sugiyama2012density}
Masashi Sugiyama, Taiji Suzuki, and Takafumi Kanamori.
\newblock {\em Density Ratio Estimation in Machine Learning}.
\newblock Cambridge University Press, UK, 2012.

\bibitem[\protect\citeauthoryear{Xu \bgroup \em et al.\egroup }{2022}]{xu2022a}
Renzhe Xu, Xingxuan Zhang, Zheyan Shen, Tong Zhang, and Peng Cui.
\newblock A theoretical analysis on independence-driven importance weighting
  for covariate-shift generalization.
\newblock In {\em Proceedings of the 39th International Conference on Machine
  Learning}, pages 24803--24829, Baltimore, USA, 2022.

\bibitem[\protect\citeauthoryear{Yoon \bgroup \em et al.\egroup
  }{2018}]{yoon2018gain}
Jinsung Yoon, James Jordon, and Mihaela van~der Schaar.
\newblock {GAIN}: Missing data imputation using generative adversarial nets.
\newblock In {\em Proceedings of the 35th International Conference on Machine
  Learning}, pages 5675--5684, Stockholmsmassan, Sweden, July 2018.

\bibitem[\protect\citeauthoryear{Zhang \bgroup \em et al.\egroup
  }{2021}]{zhang2021deep}
Xingxuan Zhang, Peng Cui, Renzhe Xu, Linjun Zhou, Yue He, and Zheyan Shen.
\newblock Deep stable learning for out-of-distribution generalization.
\newblock In {\em IEEE Conference on Computer Vision and Pattern Recognition},
  pages 5372--5382, Virtual Event, 2021.

\end{thebibliography}

\newpage
\appendix

\onecolumn

\section{Proof of Theorem \ref{thm:invariant_optimal_predictor}}\label{app:proof_invariant_optimal_predictor}

By law of total probability, the conditional label distribution given observed feature values and mask $p(\by\mid\bx_\mathbf{m},\mathbf{m})$ can be factorized as
\begin{equation*}
    \begin{aligned}
        p(\by\mid\bx_\mathbf{m},\mathbf{m}) & =\int p(\by\mid\bx_{\overline{\mathbf{m}}},\bx_{\mathbf{m}},\mathbf{m})p(\bx_{\overline{\mathbf{m}}}\mid\bx_\mathbf{m},\mathbf{m})d\bx_{\overline{\mathbf{m}}} \\
        & =\int p(\by\mid\bx,\mathbf{m})p(\bx_{\overline{\mathbf{m}}}\mid\bx_\mathbf{m},\mathbf{m})d\bx_{\overline{\mathbf{m}}} \\
        & =\int p(\by\mid\bx)p(\bx_{\overline{\mathbf{m}}}\mid\bx_\mathbf{m},\mathbf{m})d\bx_{\overline{\mathbf{m}}}\text{,}
    \end{aligned}
\end{equation*}
where in the last step we used the conditional independence $p(\by\mid\bx,\mathbf{m})=p(\by\mid\bx)$, since we consider the label generation process that does not depend on mask. The first term
\begin{equation*}
    p(\by\mid\bx)=\frac{p(\bx,\by)}{\int p(\bx,\by^\prime)d\by^\prime}
\end{equation*}
is invariant according to Assumption \ref{assump:invariant_joint_distribution}. In the case of MCAR or MAR where the missing feature $\bX_{\overline\bM}$ and mask $\bM$ are independent given the observed feature $\bX_{\bM}$,
\begin{equation*}
    p(\bx_{\overline{\mathbf{m}}}\mid\bx_{\mathbf{m}},\mathbf{m})=p(\bx_{\overline{\mathbf{m}}}\mid\bx_{\mathbf{m}})=\frac{\int p(\bx,\by)d\by}{\int p(\bx,\by)d\bx_{\overline{\mathbf{m}}}d\by}\text{,}
\end{equation*}
which is also invariant according to Assumption \ref{assump:invariant_joint_distribution}. Thus $p(\by\mid\bx_\mathbf{m},\mathbf{m})$ is invariant between training and testing.

\section{Proof of Theorem \ref{thm:unequal_testing_performance}}\label{app:proof_unequal_testing_performance}

We give the following instance claimed by Theorem \ref{thm:unequal_testing_performance}. 
Suppose $Y=X_1+2X_2$, where $X_1$ and $X_2$ are independent with Bernoulli distribution:
\begin{equation*}
p(X_i=0)=p(X_i=1)=\frac{1}{2},\quad i=1,2\text{.}
\end{equation*}
The training mask distribution is
\begin{equation*}
    p(\bM=(0,1))=p(\bM=(1,0))=\frac{1}{2}\text{,}
\end{equation*}
and $M_1$ and $M_2$ are independent with Bernoulli distribution in the testing distribution:
\begin{equation*}
    p(M_i=0)=p(M_i=1)=\frac{1}{2},\quad i=1,2\text{.}
\end{equation*}

The optimal predictor is
\begin{equation*}
    \E[Y\mid\bx_{\mathbf{m}},\mathbf{m}]=\phi_0+\sum_{i=1}^n\phi_i\cdot(\bx\odot\mathbf{m})_i\text{,}
\end{equation*}
where $\phi_0=\frac{1}{2}\overline{m}_1+\overline{m}_2$, $\phi_1=m_1$, and $\phi_2=2m_2$. Suppose we learn the optimal predictor by
\begin{equation*}
    g_{\bphi_{\btheta}(\mathbf{m})}(\bx\odot \mathbf{m})=\sum_{i=0}^2\sum_{j=0}^2\sum_{k=0}^2\theta_{ijk} m_im_jm_k x_k\text{,}
\end{equation*}
where $\btheta\in\R^{3\times3\times3}$, and we append an additional dimension of $x_0=m_0=1$ to $\bx$ and $\mathbf{m}$ to allow for the constant term. Note the parameters $\phi_k$ is given by
\[
\phi_k = \sum_{i=0}^2 \sum_{j=0}^2 \theta_{ijk} m_i m_j, \quad k=0,1,2\text{.}
\]


The parameter $\btheta^*$ corresponding to the optimal $\bphi$ is
\begin{equation*}
    \btheta_{:,:,0}^*=\begin{pmatrix}3/2&-1/2&-1\\ 0&0&0\\ 0&0&0\end{pmatrix}\text{,} 
\quad
    \btheta_{:,:,1}^*=\begin{pmatrix}0&1&0\\ 0&0&0\\ 0&0&0\end{pmatrix}\text{,}
\quad
    \btheta_{:,:,2}^*=\begin{pmatrix}0&0&2\\ 0&0&0\\ 0&0&0\end{pmatrix}\text{.}
\end{equation*}

Recall the population loss
\begin{equation*}
    \ell(\btheta)=\E_{(\bX,\bM,\bY)\sim p^{tr}}\|\bY-g_{\bphi_{\btheta}(\bM)}(\bX_\bM,\bM)\|_2^2\text{.}
\end{equation*}
It can be easily checked that the parameter $\hat\btheta$ given by
\begin{equation*}
    \hat\btheta_{:,:,0}=\begin{pmatrix}3/2&-1&-3/2\\ -1/2&1&0\\ -1/2&0&1\end{pmatrix}\text{,}
\quad
    \hat\btheta_{:,:,1}=\begin{pmatrix}0&3/2&1/2\\ 1/2&-1&0\\ 1/2&0&-1\end{pmatrix}\text{,}
\quad
    \hat\btheta_{:,:,2}=\begin{pmatrix}0&1&2\\ 1&1&1 \\ 0&1&0\end{pmatrix}
\end{equation*}
performs equally well as $\btheta^*$ on training distribution:
\begin{equation*}
    \ell_{tr}(\hat\btheta)=\ell_{tr}(\btheta^*)=\frac{5}{8}\text{.}
\end{equation*}
However, $\hat\btheta$ performs worse than $\btheta^*$ on testing distribution:
\begin{equation*}
    \ell_{te}(\hat\btheta)=\frac{15}{4}>\ell_{te}(\btheta^*)=\frac{5}{8}.
\end{equation*}

\section{Proof for Example in Section \ref{sec:prediction_framework}}\label{app:proof_decorrelation}

Note that the condition \eqref{eq:equality} in Section \ref{sec:decorrelation} is equivalent to
\begin{equation*}
    \E_{p^{tr}}\left[g_{\bphi_{\hat\btheta}(\bM)}(\bX_\bM,\bM)-g_{\bphi_{\btheta^*}(\bM)}(\bX_\bM,\bM)\right]^2=0\text{.}
\end{equation*}
The left-hand side can be expanded using Equation \eqref{eq:g} as follows:
\begin{equation*}
    \begin{aligned}
        0= & \E_{p^{tr}}[(\sum_{i,j,k}\theta_{ijk}^*X_kM_iM_jM_k-\sum_{i,j,k}\hat\theta_{ijk}X_kM_iM_jM_k)^2] \\
        = & \sum_{i,j,k}(\theta_{ijk}^*-\hat\theta_{ijk})^2\Var(X_kM_iM_jM_k)  +\left(\sum_{i,j,k}(\theta_{ijk}^*-\hat\theta_{ijk})\E[X_kM_iM_jM_k]\right)^2\\
        & +\sum_{\substack{i,j,k,s,t,l\\(i,j,k)\neq(s,t,l)}}(\theta_{ijk}^*-\hat\theta_{ijk})(\theta_{stl}^*-\theta_{stl}) \Cov(X_kM_iM_jM_k,X_lM_sM_tM_l) \text{.}
  \end{aligned}
\end{equation*}
If the entries of $\bX$ and the entries of $\bM$ are mutually independent, all the covariance terms are zero. Then the only solution for $\hat\btheta$ is the $\btheta^*$ corresponding to Equation \eqref{eq:example}.

\section{Random Fourier Feature}\label{app:rff}

Let $\cH_{\mathrm{RFF}}$ denote the function space of Random Fourier Features in the form of
\begin{equation*}
    \begin{aligned}
        \cH_{\mathrm{RFF}}=\biggl\{ & h:z\mapsto\sqrt{2}\cos(\omega z+\beta)\; \Big| \; \omega\sim\cN(0,1),\beta\sim\mathrm{Uniform}[0,2\pi)\biggr\}\text{,}
    \end{aligned}
\end{equation*}
where $\omega$ is drawn from standard Gaussian distribution, and $\beta$ is uniformly drawn from $[0,2\pi)$. Suppose we aim to measure the correlation between two random variables $A$ and $B$ with $N$ samples $\{(A^{(i)},B^{(i)})\}_{i=1}^{N}$. The partial cross-covariance matrix \citep{strobl2019approximate} of $A$ and $B$ is defined as
\begin{equation*}
    \begin{aligned}
        \hat\Sigma_{A,B}=\frac{1}{N-1}\sum_{i=1}^{N}\biggl[ & \biggl(\bu(A^{(i)})-\frac{1}{N}\sum_{j=1}^{N}\bu(A^{(j)})\biggr)^\top  \biggl(\bv(B^{(i)})-\frac{1}{N}\sum_{j=1}^{N}\bv(B^{(j)})\biggr)\biggr]\text{,}
    \end{aligned}
\end{equation*}
where
\begin{equation*}
    \begin{aligned}
        \bu(A^{(i)})=(u_1(A^{(i)}),\dots,u_q(A^{(i)})),\quad u_j(\cdot)\in\cH_{\mathrm{RFF}},\ \forall j\text{,} \\
        \bv(B^{(i)})=(v_1(B^{(i)}),\dots,v_q(B^{(i)})),\quad v_j(\cdot)\in\cH_{\mathrm{RFF}},\ \forall j\text{.}
    \end{aligned}
\end{equation*}
Functions $u_j(\cdot)$'s and $v_j(\cdot)$'s are drawn from $\cH_{\mathrm{RFF}}$ by sampling $\omega$'s and $\beta$'s from the corresponding distributions. The vectors $\bu(A^{(i)})$ and $\bv(B^{(i)})$ are used as higher dimensional representation of $A^{(i)}$ and $B^{(i)}$. The term in the summation is the cross-covariance matrix of $\bu(A^{(i)})$ and $\bv(B^{(i)})$ for the $i$-th sample, and the partial cross-covariance matrix between $A$ and $B$ is then the sum of the cross-covariance matrices for all the samples. The correlation between $A$ and $B$ is measured by the squared Frobenius norm of the partial cross-covariance matrix $\|\hat\Sigma_{AB}\|_F^2$, which is non-negative. When it decreases to zero, $A$ and $B$ become mutually independent \citep{zhang2021deep}.

Under distribution $\tilde{p}_w$, the partial cross-covariance matrix of $A$ and $B$ is given by
\begin{equation*}
    \begin{aligned}
        \hat\Sigma_{A,B;w}=\frac{1}{N-1}\sum_{i=1}^{N}\biggl[ & \biggl(w_i\bu(A^{(i)})-\frac{1}{N}\sum_{j=1}^{N}w_j\bu(A^{(j)})\biggr)^\top   \biggl(w_i\bv(B^{(i)})-\frac{1}{N}\sum_{j=1}^{N}w_j\bv(B^{(j)})\biggr)\biggr]\text{,}
    \end{aligned}
\end{equation*}
where $w_i$ is the weight for sample $(A^{(i)},B^{(i)})$. In this paper, we adapt the above matrix to the case of incomplete data.

For $X_k$ and $X_l$, $\Sigma_{X_k,X_l;w}^\prime$ is detailed in Section \ref{sec:decorrelation}. For $X_k$ and $M_l$, $\Sigma_{X_k,M_l;w}^\prime$ is computed by
\begin{equation*}
    \Sigma_{X_k,M_l;w}^\prime=\frac{1}{N^k-1}\sum_{i=1}^{N^k}\biggl[\biggl(w_i\bu(X_k^{(i)})-\frac{1}{N^k}\sum_{j=1}^{N^k}w_j\bu(X_k^{(j)})\biggr)^\top\biggl(w_i\bv(M_l^{(i)})-\frac{1}{N}\sum_{j=1}^{N}w_j\bv(M_l^{(j)})\biggr)\biggr]\text{.}
\end{equation*}
For $M_k$ and $M_l$, $\Sigma_{X_k,M_l;w}^\prime$ is computed by
\begin{equation*}
    \Sigma_{M_k,M_l;w}^\prime=\frac{1}{N-1}\sum_{i=1}^{N}\biggl[\biggl(w_i\bu(M_k^{(i)})-\frac{1}{N}\sum_{j=1}^{N}w_j\bu(M_k^{(j)})\biggr)^\top\biggl(w_i\bv(M_l^{(i)})-\frac{1}{N}\sum_{j=1}^{N}w_j\bv(M_l^{(j)})\biggr)\biggr]\text{.}
\end{equation*}

\section{Examples of Missing Patterns}\label{app:missing_pattern_example}

\begin{figure}[t]
	\centering
	\subfigure[MCAR-Ind]{
		\centering
		\includegraphics[width=0.45\columnwidth]{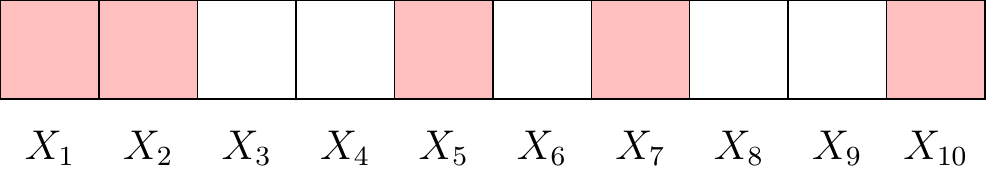}
	}
	\hspace{1ex}
	\subfigure[MCAR]{
		\centering
		\includegraphics[width=0.45\columnwidth]{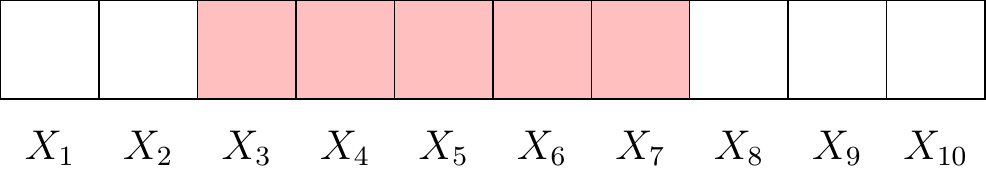}
	}
	\vspace{1ex}
	\subfigure[MAR]{
		\centering
		\includegraphics[width=0.45\columnwidth]{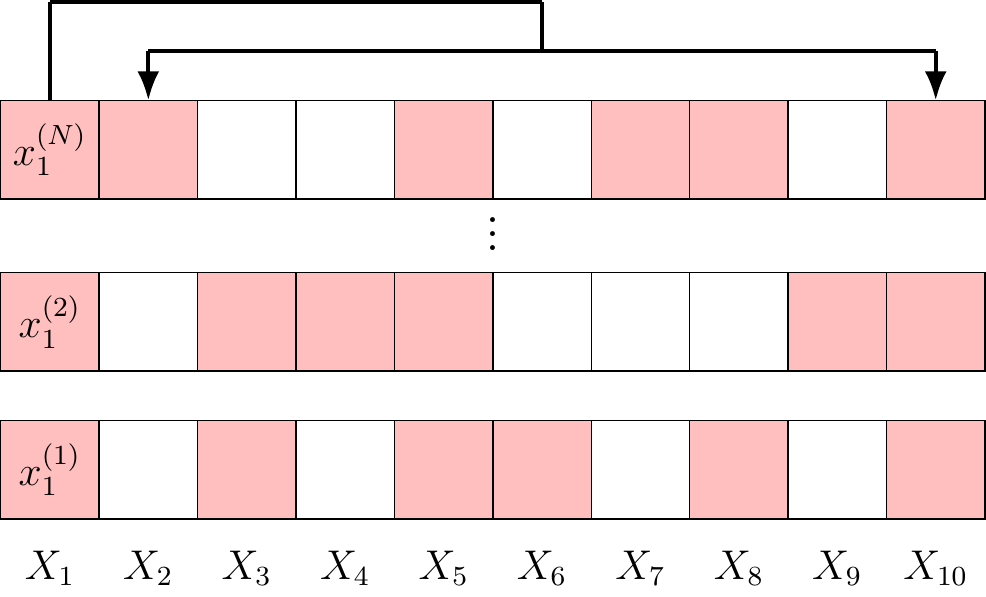}
	}
	\caption{Examples of missing patterns.}
	\label{fig:missing_pattern_example}
\end{figure}

We give examples of the missing patterns as follows, where the feature dimension $n=10$ and sample missing rate $r_s=50\%$. The examples of MCAR-Ind, MCAR and MAR are shown in Figure \ref{fig:missing_pattern_example}.

\textbf{MCAR-Ind}. The entries $X_1,\dots,X_{10}$ are mutually independently missing with probability 50\%.

\textbf{MCAR}. There is window of length $10\times50\%=5$ at a random position, in which the consecutive features are missing.

\textbf{MAR}. First, randomly selected $10\times10\%=1$ feature, namely $X_1$ in this example, is set to be observed in all samples. The mask on the other features $X_2,\dots,X_{10}$ are determined by the value of $X_1$.

\section{Implementation Details}\label{app:implementation_details}

Our model is implemented with PyTorch. The model architectures for the datasets are as follows.

\textbf{Synthetic Data}. Predictor $g=\phi_0+\sum_{i=1}^{50}\phi_i\cdot(\bx\odot\mathbf{m})_i$, and $\bphi(\mathbf{m})$ takes the form of an MLP with $n+1=51$ neurons in the input and output layers to allow for a constant term. We append an additional dimension $x_0=\mathbf{m}_0=1$ to $\bx$ and $\mathbf{m}$.

\textbf{House Sales}. Predictor $g$ takes the form of an MLP with $n+1=17$ input neurons and $1$ neuron for output. $\bphi(\mathbf{m})$ also takes the form of an MLP with $17$ input neurons. The number of output neurons is the same as that of parameters in $g$.

\textbf{MNIST}. Predictor $g$ takes the form of a $6$-layer CNN with $28\times28$ input and output feature map. $\bphi(\mathbf{m})$ is composed of a $6$-layer CNN and an MLP. The CNN takes $28\times28$ feature map as input, and the number of output neurons in the MLP is the same as that of parameters in $g$.

\textbf{Traffic}. We predict 8 future timestamps with 8 historical timestamps. The input is composed of a $1343\times8$ matrix for the historical traffic speed and a $1343\times1343$ adjacency matrix for the underlying graph. Predictor $g$ is composed of three modules: GAT, Multi-Head Self-Attention and 1-D CNN. It has an auto-encoder structure: the input first passes through a layer of GAT, a layer of Multi-Head Self-Attention and several layers of 1-D CNN, and then passes through these layers in the reverse order. The output is a $1343\times8$ matrix for the predicted future traffic speed. The architecture of $\bphi(\mathbf{m})$ is almost the same as $g$, except that the auto-encoder is followed an MLP, whose number of output neurons is the same as that of parameters in $g$.

We use Partial VAE, MIWAE and P-BiGAN for prediction by treating the label as missing feature. The size of their input and output feature maps are both the concatenation of that for feature and that for label. We use similar neural networks to $g$ in these models, except for the input and output layers for a different size of feature map. We use an improved version of NeuMiss with shared weights and residual connection \citep{morvan2021whats}. NeuMiss may be combined with an MLP as given in \citet{morvan2021whats}, but we still call it NeuMiss for short. Whether an MLP is combined is also tuned. For DWR and SRDO, we predict each label dimension separately using their linear regression. The input is flattened when fed into NeuMiss, DWR and SRDO. For StableNet, we use the same neural network as $g$. The activation function between layers is $\mathrm{ReLU}$. The number of MLP layers and hidden neurons and the kernel size are set as hyper-parameters and tuned with a validation set. The number of MLP layers is selected from $\{1,2,3\}$, the number of hidden neurons is selected from exponentials of $2$ from $[128, 4096]$, and the kernel size is selected from $\{3, 5, 7, 9, 11, 13\}$.

NeuMiss, DWR and SRDO have a small number of parameters, since they take advantage of the linear label generation process. Our model has more parameters than them, especially for the real-world datasets, as we use more complex neural networks to capture the possibly nonlinear correlation. Our model has slightly more parameters than Partial VAE, MIWAE, P-BiGAN and StableNet, mostly due to the MLP that reshapes the size of output in $\bphi(\mathbf{m})$. However, these baselines cannot increase their number of parameters in a straightforward way to improve generalizability.

For the synthetic datasets, we generate $16384$ samples for training, validation and testing, respectively; for House Sales and MNIST, we use 90\%, 5\% and 5\% of the datasets for training, validation and testing, respectively; for the naturally incomplete Traffic dataset, we use the incomplete data from Jan to Jun for training and that from Jul to Dec for testing. Hyper-parameters are tuned with the aim of minimizing the training loss. We set $q=5$, which is enough to judge the independence \citep{strobl2019approximate}. We select $\gamma$ from $\{0.1, 0.2, 0.5, 1, 2, 5, 10\}$. We use Adam optimizer with learning rate of 0.001. The models are trained for 1000 epochs with batch size of 64.

\section{Performance of Missing Rate Shift on Other Synthetic Settings}\label{app:other-performance}

The results on Gaussian-Ind feature with MCAR-Ind mask, Gaussian feature with MCAR and MAR mask, and Gaussian-Mix feature with MCAR mask when trained under 50\% missing level are shown in Table \ref{tbl:gaussian-ind_mcar-ind_fixed-training}, \ref{tbl:gaussian_mcar_fixed-training}, \ref{tbl:gaussian_mar_fixed-training} and \ref{tbl:gaussian-mix_mcar_fixed-training}, respectively.

When the mask distribution shifts, {\ours} achieves the best generalization performance on all the settings, reducing the gap to optimal of the second best, namely NeuMiss, by 6\%-10\% on Gaussian-Ind with MCAR-Ind, by 8-31\% with an average of 20\% on Gaussian with MCAR, by 23\%-39\% on Gaussian with MAR, and by 14\%-39\% with an average of 30\% on Gaussian-Mix with MCAR.

When there is no mask distribution shift, {\ours} has the best performance on Gaussian-Ind with MCAR-Ind and Gaussian-Mix with MCAR, slightly outperforming NeuMiss by 0.5\% and 6\%, respectively, in terms of the gap to optimal. NeuMiss slightly outperform {\ours} on Gaussian with MCAR or MAR, by 2\% and 0.9\%, respectively, as these are the cases that NeuMiss is tailored for. In general, {\ours} has competitive performance to NeuMiss in the absence of mask distribution shift.

\begin{table}[h]
    \centering
    \caption{Performance on Gaussian-Ind feature with MCAR-Ind mask when trained under 50\% missing level.}
    \label{tbl:gaussian-ind_mcar-ind_fixed-training}
    \begin{tabular}{@{}clrrrrrrrrr}
        \toprule
        \multicolumn{2}{c}{\multirow{2.5}*{\textbf{Method}}} & \multicolumn{9}{c}{\textbf{Testing Missing Level}} \\ \cmidrule{3-11}
        ~ & ~ & \textbf{10\%} & \textbf{20\%} & \textbf{30\%} & \textbf{40\%} & \textbf{50\%} & \textbf{60\%} & \textbf{70\%} & \textbf{80\%} & \textbf{90\%} \\
        \midrule
        \multirow{8}*{\specialcell{\textbf{Gap to}\\\textbf{Optimal}}} & Partial VAE & 1406.08 & 1140.32 & 980.69 & 895.57 & 640.24$^*$ & 762.64 & 915.90 & 899.07 & 1036.22 \\
        ~ & MIWAE & 1218.07 & 979.29 & 873.35 & 734.73 & 590.85$^*$ & 638.53 & 674.69 & 752.48 & 763.40 \\
        ~ & P-BiGAN & 1218.60 & 1064.99 & 914.26 & 784.24 & 560.52$^*$ & 621.73 & 749.38 & 806.48 & 857.55 \\
        ~ & NeuMiss & \underline{256.50} & \underline{235.24} & \underline{209.40} & \underline{172.76} & \underline{140.64}$^*$ & \underline{175.88} & \underline{195.36} & \underline{199.11} & \underline{226.28} \\
        \cmidrule{2-11}
        ~ & DWR & 1371.73 & 1158.83 & 987.23 & 883.78 & 740.86$^*$ & 762.85 & 821.43 & 902.41 & 882.14 \\
        ~ & SRDO & 1285.99 & 1045.82 & 932.55 & 818.79 & 690.25$^*$ & 738.95 & 770.21 & 790.12 & 784.29 \\
        ~ & StableNet & 1127.57 & 938.42 & 811.18 & 768.29 & 590.08$^*$ & 631.45 & 644.83 & 639.87 & 631.42 \\
        \cmidrule{2-11}
        ~ & {\ours} & \textbf{237.47} & \textbf{213.84} & \textbf{188.75} & \textbf{161.22} & \textbf{139.92}$^*$ & \textbf{160.14} & \textbf{179.74} & \textbf{184.61} & \textbf{204.27} \\
        \midrule
        \multicolumn{2}{c}{\textbf{Optimal}} & 1012.56 & 1157.22 & 1259.07 & 1372.25 & 1476.83 & 1554.6 & 1644.45 & 1724.16 & 1809.25 \\
        \bottomrule
    \end{tabular}
\end{table}


\begin{table*}[h]
    \centering
    \caption{Performance on Gaussian feature with MCAR mask when trained under 50\% missing level.}
    \label{tbl:gaussian_mcar_fixed-training}
    \begin{tabular}{@{}clrrrrrrrrr}
        \toprule
        \multicolumn{2}{c}{\multirow{2.5}*{\textbf{Method}}} & \multicolumn{9}{c}{\textbf{Testing Missing Level}} \\ \cmidrule{3-11}
        ~ & ~ & \textbf{10\%} & \textbf{20\%} & \textbf{30\%} & \textbf{40\%} & \textbf{50\%} & \textbf{60\%} & \textbf{70\%} & \textbf{80\%} & \textbf{90\%} \\
        \midrule
        \multirow{8}*{\specialcell{\textbf{Gap to}\\\textbf{Optimal}}} & Partial VAE & 820.84 & 744.93 & 641.00 & 546.41 & 269.35$^*$ & 677.19 & 755.83 & 966.18 & 1062.42 \\
        ~ & MIWAE & 624.46 & 586.45 & 547.15 & 424.09 & 219.56$^*$ & 505.89 & 621.72 & 708.05 & 762.95 \\
        ~ & P-BiGAN & 664.51 & 612.36 & 568.14 & 420.16 & 189.24$^*$ & 490.22 & 655.68 & 693.78 & 844.22 \\
        ~ & NeuMiss & \underline{386.02} & \underline{315.53} & \underline{324.83} & \underline{244.72} & \textbf{169.46}$^*$ & \underline{202.80} & \underline{288.81} & \underline{352.80} & \underline{431.05} \\
        \cmidrule{2-11}
        ~ & DWR & 769.62 & 756.13 & 692.83 & 572.36 & 369.26$^*$ & 703.12 & 783.65 & 846.27 & 918.03 \\
        ~ & SRDO & 696.75 & 703.81 & 636.75 & 532.20 & 319.10$^*$ & 630.28 & 697.67 & 734.57 & 822.00 \\
        ~ & StableNet & 562.25 & 587.86 & 555.83 & 420.75 & 219.53$^*$ & 505.09 & 548.23 & 584.96 & 663.27 \\
        \cmidrule{2-11}
        ~ & {\ours} & \textbf{302.99} & \textbf{297.97} & \textbf{260.82} & \textbf{190.70} & \underline{172.08}$^*$ & \textbf{187.58} & \textbf{202.67} & \textbf{268.30} & \textbf{296.04} \\
        \midrule
        \multicolumn{2}{c}{\textbf{Optimal}} & 516.26 & 547.62 & 582.35 & 642.66 & 745.85 & 852.35 & 970.87 & 1093.49 & 1224.33 \\
        \bottomrule
    \end{tabular}
\end{table*}


\begin{table*}[h]
    \centering
    \caption{Performance on Gaussian feature with MAR mask when trained under 50\% missing level.}
    \label{tbl:gaussian_mar_fixed-training}
    \begin{tabular}{@{}clrrrrrrrrr}
        \toprule
        \multicolumn{2}{c}{\multirow{2.5}*{\textbf{Method}}} & \multicolumn{9}{c}{\textbf{Testing Missing Level}} \\ \cmidrule{3-11}
        ~ & ~ & \textbf{10\%} & \textbf{20\%} & \textbf{30\%} & \textbf{40\%} & \textbf{50\%} & \textbf{60\%} & \textbf{70\%} & \textbf{80\%} & \textbf{90\%} \\
        \midrule
        \multirow{8}*{\specialcell{\textbf{Gap to}\\\textbf{Optimal}}} & Partial VAE & 1054.20 & 1055.49 & 835.67 & 734.95 & 621.92$^*$ & 906.54 & 806.07 & 877.27 & 926.30 \\
        ~ & MIWAE & 770.85 & 787.16 & 753.76 & 669.59 & 571.34$^*$ & 718.15 & 682.46 & 672.22 & 735.83 \\
        ~ & P-BiGAN & 890.55 & 815.86 & 727.53 & 710.58 & 541.75$^*$ & 679.64 & 671.41 & 747.59 & 752.90 \\
        ~ & NeuMiss & \underline{568.27} & \underline{539.71} & \underline{447.97} & \underline{370.06} & \textbf{219.42}$^*$ & \underline{393.40} & \underline{477.20} & \underline{509.01} & \underline{538.29} \\
        \cmidrule{2-11}
        ~ & DWR & 930.13 & 938.99 & 873.49 & 853.3 & 721.35$^*$ & 861.01 & 798.41 & 805.55 & 809.68 \\
        ~ & SRDO & 844.58 & 870.45 & 832.96 & 754.17 & 671.72$^*$ & 855.99 & 721.94 & 779.74 & 768.08 \\
        ~ & StableNet & 720.77 & 737.80 & 690.29 & 658.02 & 571.57$^*$ & 704.95 & 603.50 & 592.94 & 637.19 \\
        \cmidrule{2-11}
        ~ & {\ours} & \textbf{435.24} & \textbf{353.15} & \textbf{333.96} & \textbf{266.04} & \underline{221.36}$^*$ & \textbf{268.18} & \textbf{288.06} & \textbf{324.51} & \textbf{376.28} \\
        \midrule
        \multicolumn{2}{c}{\textbf{Optimal}} & 546.37 & 551.64 & 574.84 & 633.26 & 742.58 & 776.75 & 952.50 & 965.95 & 1140.43 \\
        \bottomrule
    \end{tabular}
\end{table*}


\begin{table*}[h]
    \centering
    \caption{Performance on Gaussian-Mix feature with MCAR mask when trained under 50\% missing level.}
    \label{tbl:gaussian-mix_mcar_fixed-training}
    \begin{tabular}{@{}clrrrrrrrrr}
        \toprule
        \multicolumn{2}{c}{\multirow{2.5}*{\textbf{Method}}} & \multicolumn{9}{c}{\textbf{Testing Missing Level}} \\ \cmidrule{3-11}
        ~ & ~ & \textbf{10\%} & \textbf{20\%} & \textbf{30\%} & \textbf{40\%} & \textbf{50\%} & \textbf{60\%} & \textbf{70\%} & \textbf{80\%} & \textbf{90\%} \\
        \midrule
        \multirow{8}*{\specialcell{\textbf{Gap to}\\\textbf{Optimal}}} & Partial VAE & 1370.80 & 1138.15 & 1022.81 & 925.79 & 872.67$^*$ & 916.44 & 1058.23 & 1263.68 & 1299.88 \\
        ~ & MIWAE & 1141.75 & 1006.24 & 977.36 & 900.11 & 825.64$^*$ & 906.45 & 953.03 & 1035.04 & 1019.08 \\
        ~ & P-BiGAN & 1249.11 & 991.07 & 948.68 & 864.18 & 794.86$^*$ & 864.54 & 942.71 & 1125.61 & 1336.18 \\
        ~ & NeuMiss & \underline{607.77} & \underline{567.99} & \underline{480.56} & \underline{373.15} & \underline{312.75}$^*$ & \underline{478.63} & \underline{527.64} & \underline{544.12} & \underline{665.30} \\
        \cmidrule{2-11}
        ~ & DWR & 1290.18 & 1070.35 & 931.87 & 928.45 & 873.53$^*$ & 944.86 & 1072.58 & 1223.20 & 1132.42 \\
        ~ & SRDO & 1187.14 & 1086.90 & 932.88 & 863.52 & 826.52$^*$ & 873.37 & 1021.75 & 1172.92 & 1217.66 \\
        ~ & StableNet & 1061.32 & 906.95 & 833.19 & 779.37 & 725.82 $^*$& 807.52 & 909.54 & 991.59 & 1190.70 \\
        \cmidrule{2-11}
        ~ & {\ours} & \textbf{409.74} & \textbf{390.43} & \textbf{346.55} & \textbf{319.13} & \textbf{292.11}$^*$ & \textbf{323.41} & \textbf{331.50} & \textbf{359.62} & \textbf{403.29} \\
        \midrule
        \multicolumn{2}{c}{\textbf{Optimal}} & 861.06 & 1035.55 & 1134.42 & 1206.60 & 1301.95 & 1411.94 & 1591.95 & 1673.51 & 1795.19 \\
        \bottomrule
    \end{tabular}
\end{table*}


\clearpage

\section{Performance of Missing Pattern Shift}\label{app:pattern_shift}

The results of missing pattern shift on Gaussian-Mix feature from MCAR to MAR is shown in Table \ref{tbl:gaussian-mix_mcar_to_mar} and from MAR to MCAR is shown in Table \ref{tbl:gaussian-mix_mar_to_mcar}. {\ours} achieves the best generalization performance on all the settings, reducing the gap to optimal of the second best, namely NeuMiss, by 39\% from MCAR to MAR and by 24\% from MAR to MCAR.

\begin{table*}[h]
    \centering
    \caption{Performance on Gaussian-Mix feature when trained with MCAR and tested with MAR under corresponding missing rates.}
    \label{tbl:gaussian-mix_mcar_to_mar}
    \begin{tabular}{@{}clrrrrrrrrr}
        \toprule
        \multicolumn{2}{c}{\multirow{2.5}*{\textbf{Method}}} & \multicolumn{9}{c}{\textbf{Missing Level}} \\ \cmidrule{3-11}
        ~ & ~ & \textbf{10\%} & \textbf{20\%} & \textbf{30\%} & \textbf{40\%} & \textbf{50\%} & \textbf{60\%} & \textbf{70\%} & \textbf{80\%} & \textbf{90\%} \\
        \midrule
        \multirow{8}*{\specialcell{\textbf{Gap to}\\\textbf{Optimal}}} & Partial VAE & 1493.39 & 1589.29 & 1628.40 & 1687.72 & 1736.49 & 1812.35 & 1883.28 & 1938.03 & 1985.35 \\
        ~ & MIWAE & 1257.20 & 1321.28 & 1368.02 & 1425.75 & 1536.43 & 1603.48 & 1665.92 & 1713.42 & 1788.90 \\
        ~ & P-BiGAN & 1285.82 & 1342.39 & 1390.98 & 1452.38 & 1510.20 & 1578.19 & 1616.48 & 1670.64 & 1721.14 \\
        ~ & NeuMiss & \underline{656.39} & \underline{694.28} & \underline{682.20} & \underline{732.83} & \underline{753.02} & \underline{770.92} & \underline{791.02} & \underline{824.59} & \underline{853.39} \\
        \cmidrule{2-11}
        ~ & DWR & 1435.63 & 1499.87 & 1536.09 & 1578.32 & 1623.49 & 1689.20 & 1737.62 & 1780.24 & 1815.25 \\
        ~ & SRDO & 1325.68 & 1397.06 & 1428.48 & 1496.28 & 1530.10 & 1593.82 & 1623.46 & 1673.39 & 1721.73 \\
        ~ & StableNet & 1180.37 & 1235.83 & 1289.30 & 1342.29 & 1381.20 & 1435.93 & 1482.47 & 1519.80 & 1531.25 \\
        \cmidrule{2-11}
        ~ & {\ours} & \textbf{397.92} & \textbf{401.84} & \textbf{425.50} & \textbf{442.31} & \textbf{451.24} & \textbf{475.63} & \textbf{478.29} & \textbf{499.01} & \textbf{522.46} \\
        \midrule
        \multicolumn{2}{c}{\textbf{Optimal}} & 904.87 & 1096.36 & 1108.73 & 1224.50 & 1324.00 & 1375.16 & 1518.59 & 1626.74 & 1705.54 \\
        \bottomrule
    \end{tabular}
\end{table*}


\begin{table*}[h!]
    \centering
    \caption{Performance on Gaussian-Mix feature when trained with MAR and tested with MCAR under corresponding missing rates.}
    \label{tbl:gaussian-mix_mar_to_mcar}
    \begin{tabular}{@{}clrrrrrrrrr}
        \toprule
        \multicolumn{2}{c}{\multirow{2.5}*{\textbf{Method}}} & \multicolumn{9}{c}{\textbf{Testing Missing Level}} \\ \cmidrule{3-11}
        ~ & ~ & \textbf{10\%} & \textbf{20\%} & \textbf{30\%} & \textbf{40\%} & \textbf{50\%} & \textbf{60\%} & \textbf{70\%} & \textbf{80\%} & \textbf{90\%} \\
        \midrule
        \multirow{8}*{\specialcell{\textbf{Gap to}\\\textbf{Optimal}}} & Partial VAE & 1289.37 & 1335.93 & 1384.28 & 1422.19 & 1479.90 & 1521.12 & 1556.35 & 1593.18 & 1617.33 \\
        ~ & MIWAE & 1058.23 & 1099.31 & 1146.74 & 1189.57 & 1231.46 & 1283.62 & 1327.38 & 1349.90 & 1386.82 \\
        ~ & P-BiGAN & 1170.34 & 1235.72 & 1287.48 & 1324.62 & 1373.28 & 1410.55 & 1452.85 & 1497.52 & 1527.38 \\
        ~ & NeuMiss & \underline{538.90} & \underline{570.27} & \underline{605.38} & \underline{638.16} & \underline{667.24} & \underline{695.20} & \underline{731.21} & \underline{765.27} & \underline{795.10} \\
        \cmidrule{2-11}
        ~ & DWR & 1248.26 & 1289.51 & 1336.67 & 1386.27 & 1430.61 & 1481.25 & 1539.67 & 1580.50 & 1607.31 \\
        ~ & SRDO & 1173.44 & 1213.48 & 1257.62 & 1295.47 & 1327.39 & 1371.90 & 1408.41 & 1448.89 & 1489.72 \\
        ~ & StableNet & 1095.23 & 1135.62 & 1162.84 & 1199.22 & 1259.13 & 1286.39 & 1342.17 & 1382.53 & 1437.69 \\
        \cmidrule{2-11}
        ~ & {\ours} & \textbf{384.29} & \textbf{419.26} & \textbf{445.60} & \textbf{473.75} & \textbf{498.13} & \textbf{523.69} & \textbf{561.58} & \textbf{593.45} & \textbf{610.33} \\
        \midrule
        \multicolumn{2}{c}{\textbf{Optimal}} & 861.06 & 1035.55 & 1134.42 & 1206.60 & 1301.95 & 1411.94 & 1591.95 & 1673.51 & 1795.19 \\
        \bottomrule
    \end{tabular}
\end{table*}


\end{document}